\documentclass[preprint,12pt]{elsarticle}

\usepackage{amsmath,amssymb,amsfonts}
\usepackage{algorithmic}
\usepackage{graphicx}
\usepackage{textcomp}
\usepackage{hyperref}
\usepackage[]{algorithm2e}
\SetKwRepeat{Do}{do}{while}%

\usepackage[table,xcdraw]{xcolor}
\def\BibTeX{{\rm B\kern-.05em{\sc i\kern-.025em b}\kern-.08em
    T\kern-.1667em\lower.7ex\hbox{E}\kern-.125emX}}
\usepackage{lineno}

\journal{Knowledge-Based Systems}

\begin{document}

\begin{frontmatter}



\title{Knowledge-aware equation discovery with automated background knowledge extraction}


\author[ITMO]{Elizaveta Ivanchik}
\ead{eaivanchik@itmo.ru}

\author[ITMO]{Alexander Hvatov}
\ead{alex\_hvatov@itmo.ru}


\affiliation[ITMO]{organization={ITMO University},%Department and Organization
            addressline={Kronverkskiy pr. 49A}, 
            city={Saint-Petersburg},
            postcode={197101}, 
            country={Russia}}

\begin{abstract}
In differential equation discovery algorithms, a priori expert knowledge is mainly used implicitly to constrain the form of the expected equation, making it impossible for the algorithm to truly discover equations. Instead, most differential equation discovery algorithms try to recover the coefficients for a known structure. In this paper, we describe an algorithm that allows the discovery of unknown equations using automatically or manually extracted background knowledge. Instead of imposing rigid constraints, we modify the structure space so that certain terms are likely to appear within the crossover and mutation operators. In this way, we mimic expertly chosen terms while preserving the possibility of obtaining any equation form. The paper shows that the extraction and use of knowledge allows it to outperform the SINDy algorithm in terms of search stability and robustness. Synthetic examples are given for Burgers, wave, and Korteweg--De Vries equations.

\end{abstract}

%\includegraphics[scale=0.46]{grabs}
% scale=0.7 не помещается


\begin{keyword}
equation discovery, SINDy, EPDE, SymNet, knowledge extraction, knowledge-aware algorithm, evolutionary optimization, physics-informed machine learning
\end{keyword}

\end{frontmatter}

%\linenumbers

\section{Introduction}

Knowledge extraction \cite{gomez2007semantic} and further use is a topic of many publications. Knowledge is extracted and used in the form of knowledge graphs \cite{deng2022low} and other forms \cite{gangemi2016identifying}, which are further used to improve the quality of the predictions from machine learning \cite{lan2022transfer,li2023experiencer} and other models \cite{yu2020knowledge}.

Differential equation discovery also may be considered a knowledge extraction tool in the form of a physical model, which in most cases is represented by a differential equation. The discovery of equations has deep roots \cite{langley1987heuristics} with a significant breakthrough made by SINDy \cite{brunton2016discovering} and PDE-FIND \cite{rudy2017data}. The group of equation discovery methods allows one to extract interpretable models for physical data. In this pursuit, a variety of techniques have been developed, including SINDy add-ons such as weak \cite{messenger2021weak} and ensemble methods \cite{fasel2022ensemble}, as well as other frameworks such as PDE-Net \cite{long2019pde}, DLGA-PDE \cite{xu2020dlga}, SGA-PDE, PDE-READ, EPDE \cite{maslyaev2021partial} for ordinary (ODE) and partial differential equations (PDE). In addition, many methods of symbolic regression allow one to restore an expression without differentials \cite{cornelio2023combining}.

Although we can extract knowledge in the form of an equation, there is a limited possibility to embed the background knowledge into the optimization algorithm to increase the quality of a resulting model. 
{ If we move to the algebraic (meaning that no differentials are considered within the expression), there is a lot of effort to embed the background knowledge in various forms.  

First attempts were made with expertly done grammatics \cite{bridewell2008inductive, mckay2010grammar,brence2021probabilistic}. The former also uses the first derivative as an expression language element for some cases. There is also a method for automated knowledge extraction using the Bayesian approach \cite{guimera2020bayesian}. One could argue that for the differential equation it might be harder to extract grammatical rules due to the specificity of the model. That is, we cannot extract the function (a solution) directly from the equation without a solver. Moreover, we could not evaluate the importance of each term in the equation, since we also need to solve the reduced equations. Such arguments also apply to the Bayesian approach. The approach to extract the joint distribution of the differential equation terms with a Bayesian network and a differential equation solver is described in \cite{hvatov2023towards}.
}

{Moving to the differential equation, we are not just adding differentiation operations to the expression. The differential equation determines an implicit function that could only be extracted with a solver. Therefore, we could not compute the function directly at the given grid point.} Classical { differential equation discovery algorithms use LASSO regression based on gradient optimization. The only way to use background knowledge in the regression-based equation discovery algorithm (including neural networks) is to constrain the form of the equation \cite{brunton2016discovering}. Such an approach reduces the possibility of extracting new knowledge by reducing the problem to recovering the coefficients in a known equation. Evolutionary optimization methods have more parameters to control the discovery process. }

{
Recent advances in replacing symbolic regression with reinforcement learning for differential equation discovery DISCOVER \cite{du2024discover} and \cite{zhang2024morl4pdes} could also benefit from reinforcement learning methods with automated symbloc grammar extraction \cite{crochepierre2022reinforcement} or with prior distribution \cite{landajuela2021discovering}. We note that it also includes an automatically extracted distribution process described in the paper.
}

Previously, we introduced a version of the EPDE algorithm \cite{ivanchik2023directed} that could be guided by the probability of selecting terms of the equation in the mutation and cross-over operators. We manually make the probability distribution in the reference based on the form of the equation.

The current paper is devoted to the following questions: 

\begin{itemize}
    \item What is the best form of background knowledge representation for differential equation discovery?
    \item Could we automatically extract background knowledge for differential equation discovery from data?
\end{itemize}

As a positive answer, the paper describes the complete process of background knowledge extraction in the form of a term importance distribution using an initial guess obtained by a simpler algorithm. In terms of noise robustness, overall quality, and possible equation forms, the algorithm outperforms existing methods such as different versions of SINDy (pySINDy \cite{desilva2020} is used for comparison). As mentioned above, the disadvantage of all evolutionary methods is the speed of optimization. However, the set of possible equations is much larger, which is important for the discovery of unknown equations.

The paper is organized as follows. Section \ref{sec:background} contains the statement of the equation discovery problem from a knowledge extraction point of view. Section \ref{sec:modified_alg} contains the main contribution of the paper, a knowledge-aware version of the EPDE algorithm, and a method for autonomous knowledge extraction. Section \ref{sec:experiments} contains an experimental comparison with pySINDy and the classical EPDE algorithm. Section \ref{sec:conclusion} concludes the paper.

\section{Equation discovery background}
\label{sec:background}

{ In what follows, we will discuss only differential equation discovery, and equation discovery means differential equation discovery for brevity.}

In all cases for the equation discovery problem, it is assumed that the data is placed on a discrete grid $X=\{x^{(i)}=(x^{(i)}_1,...x^{(i)}_{\text{dim}})\}_{i=1}^{i=N}$, where $N$ is the number of observations and $\text{dim}$ is the dimensionality of the problem. We mention a particular case of time series, for which $\text{dim}=1$ and $X=\{t_j\}_{i=1}^{i=N}$.

It is also assumed that for each point on the grid, there is an associated set of observations $U=\{u^{(i)}=(u^{(i)}_1,...,u^{(i)}_L)\}_{i=1}^{N}$ to define a grid map $u: X \subset \mathbb{R}^{\text{dim}} \to U \subset \mathbb{R}^L$. It is assumed that $u$ is defined by the model $M$ which has the form:

\begin{equation}
    M(S,P,x) \to u(x) : M(S,P,x^{(i)}) \to u(x_i) \sim u^{(i)}
    \label{eq:model}
\end{equation}

As discussed before, the differntial equation defines implicit function. We show it in Eq.~\ref{eq:model} with $\to$ sign. In Eq.\ref{eq:model}, we define two parts of the model in the form of the equation: the structure $S$ and the parameters $P$. We note that we do not expect either interpolation (case $M(S,P,x^{(i)}) \to u(x_i) = u^{(i)}$) or approximation case (case $M(S,P,x^{(i)}) \to u(x_i) \approx u^{(i)}$). It is assumed that the model $M(S,P,x)$ by itself may be interpreted by an expert and used, for example, to predict the behavior of the system in states that have not yet been observed $\tilde{x}^{(j)}$. 
In an ideal scenario, the discovery of differential equations enables the extraction of the complete set of underlying equations based on observational data. Unfortunately, in practical situations, we can only approximate the system and obtain a rough estimate.

The relation between grid $X$ and observation $U$ is defined by the problem statement. The structure $S$ is a computational graph of a model; it could be either a classical computational graph or a simple parametrized symbolic string-like expression. It is convenient to separate numerical characteristics such as the coefficient of the term, the power of the term, and the order of the derivative into a set of parameters $P$, i.e., make every node or element in the structure parametrized. The optimization process may be separated for structure $S$, and parameter set $P$.

In most cases, the search space of all possible structures $\Sigma$ cannot be fully explored due to its size, and apart from the extraction of knowledge from the model, we have to use a priori background knowledge (application area, modeling expertise) to reduce the set of possible structures to $\Sigma' \subset \Sigma$ and set of the possible parameters to $P' \subset P$. As a result, all methods of equation discovery differ not only with the model type (ODE, PDE, other types of expressions) but also with how the background knowledge could be incorporated into the discovery process.

There are two limiting cases of differential equation discovery methods that use background knowledge:

\begin{enumerate}
    \item[(I)] The application of gradient optimization to a fixed number of terms in a fixed structure ($u_t=F(u,u_x,u_{xx},...)$). In this case, $\Sigma'$ is a known fixed structure; it could be a weighted sum, pre-defined terms, or a neural network, as done in PDE-Net or NeuralODE. The parameter set $P'$ is reduced to the weights in a LASSO regression or a neural network. In this group of methods, the optimization is performed only in a parameter space $P'$.
    \item[(II)] The application of genetic programming to construct a computation graph using basic operations: differentiation, sum, multiplication, and power. In this case, $\Sigma'$ is a subset of $\Sigma$ of all models that could be expressed with a defined basic operation, and the space of parameters is empty, i.e., $P' =\{\varnothing\}$.
\end{enumerate}

In case (I), new information is generated in form of the coefficients $P'$, which could be the material or geometry parameters. Note that in this case we significantly reduce the ability to get new equations. In most applications, it is a coefficient restoration rather than the equation discovery. As an advantage, the search space is numeric, and thus, the optimization process is faster than in any other type of equation discovery algorithm. To make the search faster and more stable, we add background knowledge to reduce the pre-defined structure $\Sigma'$ and restrict the parameter values in $P'$ using expert knowledge of the process and presumably material or geometry parameters. We note that adding expert background knowledge to the algorithm further reduces the possibility of discovering principally new equations; for example, other process scales are left out of the scope.

We note that an approach such as PDE-READ \cite{stephany2022pde} extracts new information in form of the compact equation from the obtained within the sparse regression. It is a generalization of sequential reduction in the size of the term library.

Case (II), on contrary, allows one to get as many equation structures as possible, making the optimization process a combinatorial search that significantly increases the optimization time. In its pure form, such methods are inapplicable to practical tasks. Thus, we must use background knowledge in the form of restrictions to a possible set of structures $\Sigma'$ and, in some cases, make the nodes parametric to exchange part of the structural set $\Sigma'$ to the set of parameters $P'$. Returning to the new information, we note that we can obtain any form of the equation possible to reach as a combination of basic operations. 

Whereas case (I) appears in methods origin from SINDy \cite{messenger2021weak,Kaptanoglu2022,desilva2020} that have LASSO regression as an optimization base and others such as PDE-Net \cite{long2019pde}, it is hard to find pure case (II) and most of the alternative algorithms (from SINDy) contain a combination of structural and parametric search. {We note the PySR framework \cite{cranmer2023interpretable} as a pure (II) case for algebraic equations (by algebraic we mean that differentials are not included).}

As an example of not pure case (II), we may mention SGA-PDE and DLGA-PDE, where nodes-terms in the computational graph are represented by another graph in the SGA-PDE case or with a neural network in the DLGA-PDE case. Instead of pure structural optimization in SGA-PDE, we factor structure subspace $\Sigma'=\Sigma'_{equation} \times \Sigma'_{term}$, and thus the optimization process into equation structure optimization in space $\Sigma'_{equation}$ and term structure optimization $\Sigma'_{term}$. Both spaces $\Sigma'_{equation}$ and $\Sigma'_{term}$ have a smaller dimension than $\Sigma'$, and thus the optimization process is more computationally effective. The main difference in the DLGA-PDE approach is that $\Sigma'_{term}$ uses a neural network and is mapped in a pure numerical optimization of neural network training to $P'_{term}$.

We mention that the backgorund knowledge is mainly used to organize the search space, i.e., the model representation. A priori area knowledge, as in case (I), could only restrict the set of possible structures in this case. Namely, we expect a specific equation and try to restrain the search space such that it is obtained in most of the runs. As an advantage in case based on (II), the expert gets more tuning parameters so that the structure restrictions could be softened and a theoretically broader class of equations may be obtained even if the possible structure set $\Sigma'$ is reduced such that a specific equation is expected. 

In both cases, data-driven information extraction procedure loops up. We must expect a particular equation and convey it to the equation discovery algorithm to restore it -- we cannot get a new equation. Therefore, we require a tool to retrieve and convey background knowledge to an algorithm with fewer structural and parametric restrictions.

\subsection{Classical evolutionary equation discovery algorithm}
\label{sec:classics}

As a ground, we select the EPDE differential discovery algorithm based on evolutionary optimization. As an optimization ground, it uses rather memetic algorithm. This section briefly describes the chromosome and operators used within the optimization.

\paragraph{Model definition} Evolutionary algorithms use elementary operations to build a model structure. To reduce the amount of structural optimization, EPDE operates with building blocks -- tokens -- that are parametrized families of functions and operators. The token generally has the form shown in Eq.~\ref{eq:token}.

\begin{equation}
    t=t(\pi_1,...\pi_n)
    \label{eq:token}
\end{equation}

In Eq.~\ref{eq:token} $\pi_1,...\pi_n$ are the token parameters, explained below. In order to distinguish between a single token and a token product (term), we use the notation $T=t_1 \cdot...\cdot t_{T_{length}}$, where $0 < T_{length} \le T_{max}$, and $T_{max}$ is considered an algorithm hyperparameter. However, it is essential to note that while $T_{max}$ affects the model's final form, a reasonable value of tokens in a term (usually 2 or 3) is sufficient to capture most of the actual differential equations. 

Tokens $t_i$ are grouped into token families $\Phi_j$ to aid in fine-tuning the model form. Tokens in each family have fixed parameters set $\pi_1,...\pi_n$. For example, we could define the differential operators family $\Phi_{der}=\{\frac{\partial^{\pi_{n+1}} u}{\partial^{\pi_1}x_1 ... \partial^{\pi_n}x_n}\}$ to find linear or nonlinear equations with constant coefficients. We could also consider the trigonometric token family $\Phi_{trig}=\{\sin{(\pi_1 x_1+...+\pi_n x_n)}, \\ \cos{(\pi_1 x_1 + ...+\pi_n x_n)}\}$ to search for forcing functions or variable coefficients. 

The parameters in tokens may be optimizable and non-optimizable. For example, it is convenient to fix the differential operator parameters every time they appear; therefore, they are considered one family but different tokens and may appear multiple times in the term to reflect nonlinearity. Trigonometric tokens are optimized and appear (if required) only once per term. The algorithm takes as input the set $\Phi=\bigcup \limits_j \Phi_j$ of chosen or user-defined token families.

For simplicity, we assume the tokens are pre-computed on a discrete grid, but the grid choice does not impact the algorithm's description. Therefore, the structure and parameters are only essential for the differential equation model learning, and the model has the form Eq.~\ref{eq:diffeqn_model}.

\begin{equation}
    M(S, \{C, P\})=\sum \limits_{j=1}^{j \le N_{terms}} C_j T_j
    \label{eq:diffeqn_model}
\end{equation}

In Eq.~\ref{eq:diffeqn_model} with the structure $S$ we denote terms set $\{T_j\}_{j=1}^{j=N_{terms}}$ consisting of different tokens, the set of parameters is divided into terms coefficients $C=\{C_j\}_{j=1}^{j=N_{terms}}$ with $C_j$  and set of the optimizeable parameters $P=\{\pi_1,...\}$ without fixed length. Every model could have a different set of optimizeable parameters and the evolutionary operators could also change this number. 

The maximum number of terms $N_{terms}$ is the hyperparameter of the algorithm. We note that the hyperparameter $N_{terms}$ also has not a directive but a restrictive function. The number of terms in the resulting model may be lower than $N_{terms}$ and is eventually reduced with the fitness calculation procedure described below.

\begin{figure}[ht!]
    \centering
    \includegraphics[width=1\linewidth]{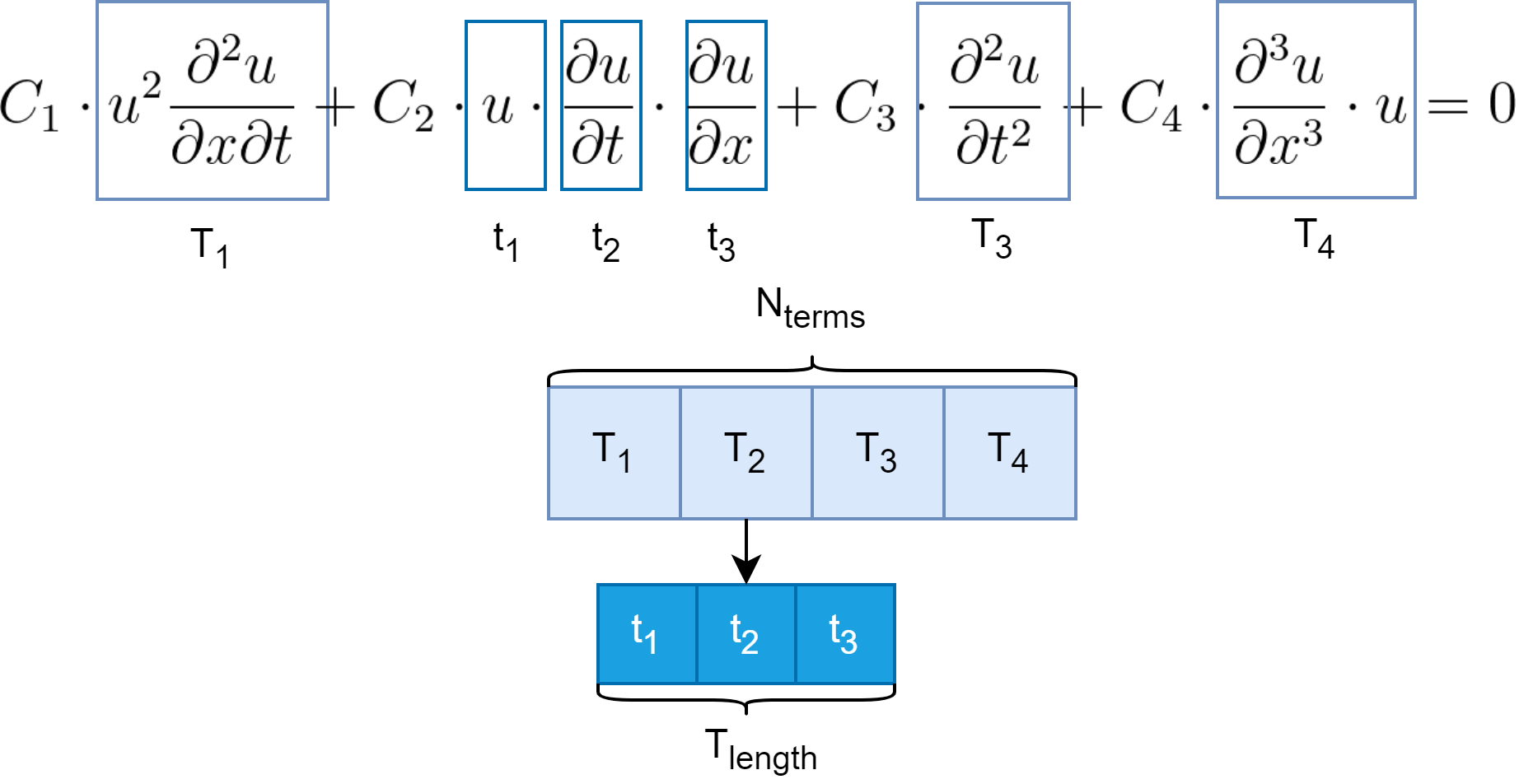}
    \caption{Model visualization: $T_i$ are the token products from Eq.~\ref{eq:model} and $t_i$ are the tokens from Eq.~\ref{eq:token}.}
    \label{fig:model_scheme}
\end{figure}

We use the simplified individual to visualize the following evolutionary operator schemes, as illustrated in Fig.~\ref{fig:model_scheme}. Each individual corresponds to an instance of the model shown in Eq.~\ref{eq:model}.

The optimization is separated into two steps: structural and parametric. The population is initialized with the models with a separate structure. After the initialization, the parametric optimization step calculates each individual's fitness.

\paragraph{Fitness evaluation}

Fitness evaluation has two purposes. First, it allows determining the parameters $\{C,P\}$ for every model -- individual. Second, it serves as a standard measure of individual fitness. To evaluate the fitness function, a term is randomly chosen as a "target" for a given model with the structure $S$. Namely, the model--individ before fitness computation transferred to the form Eq.~\ref{eq:target_meaning}.

\begin{equation}
T_{target}= \sum \limits_{j=1,...{target}-1}^{j={target}+1,...,N_{terms}} C_j T_j 
    \label{eq:target_meaning}
\end{equation}

In Eq.~\ref{eq:target_meaning} $target$ is a randomly chosen index. The randomly chosen target allows one to avoid a trivial solution $\forall j \; C_j=0$. It is assumed that for the fitness computation, the terms $T_j$ are fixed and the coefficients $C = C_1,... C_{target},... C_{N_{terms}}\}$ and optimizable parameters $P=\{P_1,...P_{target}, \\ ...P_{N_{terms}}\}$ (if any) should be determined. We note that $C_{target} \equiv -1$ and the values in the set of term parameters $P_{target}$ are always fixed.

The term coefficients $C_{opt}$ and the parameter sets $P_{opt}$ are found using LASSO regression as shown in Eq.~\ref{eq:fitness_optimization}.

\begin{multline}
        C_{opt},P_{opt}=\text{arg} \min \limits_{C,P} \Big|\Big| T_{target}-\sum \limits_{j=1,...{target}-1}^{j={target}+1,...,N_{terms}} C_j T_j \Big|\Big|_2 +\\
        + \lambda (|| C ||_1+|| P ||_1)  \}
        \label{eq:fitness_optimization}
\end{multline}

In Eq.~\ref{eq:fitness_optimization}, with $||\cdot||_p$ corresponding $l_p$ norm is designated.  After applying the LASSO regression operator, the coefficients are compared with the minimal coefficient value threshold of the term. If the absolute value of coefficient $C_j$ is lower than the threshold, then the term is removed from the current model. Thus, the model is refined to reduce the excessive growth of unnecessary terms.

After finding the final set of optimal coefficients for Eq.~\ref{eq:fitness_optimization} is found, the fitness function $F$ is calculated as shown in Eq.~\ref{eq:fitness_computation}.

\begin{equation}
F= \frac{1}{\Big|\Big| M(S,\{P_{opt},C_{opt}\}) \Big|_X \Big|\Big|_2}  
    \label{eq:fitness_computation}
\end{equation}

{ In Eq.~\ref{eq:fitness_computation} denominator is basically an average discrepancy over a computation grid $X$.}

\paragraph{Evolutionary operators}

Population initialization, cross-over, and mutation operators use a set of expert rules for generation and exchange. The rules are used to avoid situations $0=0$ (for example, two terms obtained using the commutative multiplication property are restricted to appear) or to appear of two equal terms during the mutation and cross-over steps. The rules do not change the set of obtainable equations, but they serve to avoid non-correct equations. Every structure $S_{ind}$ in model $M(S_{ind},\{P_{ind},C_{ind}\})$ has its own set of restricted tokens that could not be added to the model during cross-over and mutation operations.

Apart from the rules restrictions, all tokens in the classical algorithm may appear equiprobably during the mutation, and every term may be equiprobably exchanged during the cross-over.

\begin{figure}[ht!]
    \centering
    \includegraphics[width=0.9\linewidth]{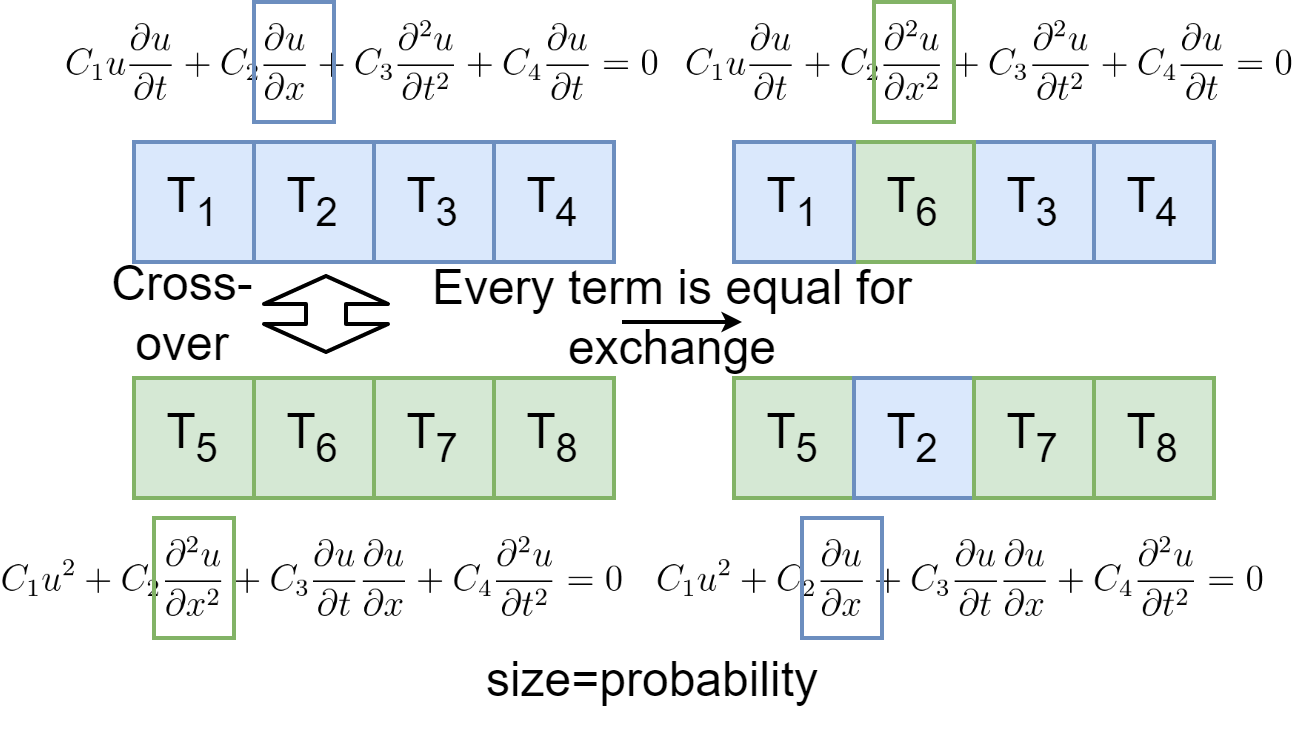}
    \caption{The classical algorithm cross-over. All terms have the same probability of participating in the cross-over.}
    \label{fig:uniform_cross_over}
\end{figure}

The cross-over operator is defined as an exchange of terms between individuals, as shown in Fig.~\ref{fig:uniform_cross_over}. We note that the terms for exchange are chosen using a uniform distribution, i.e., all terms have the same possibility of participating in the exchange.

\begin{figure}[ht!]
    \centering
    \includegraphics[width=0.9\linewidth]{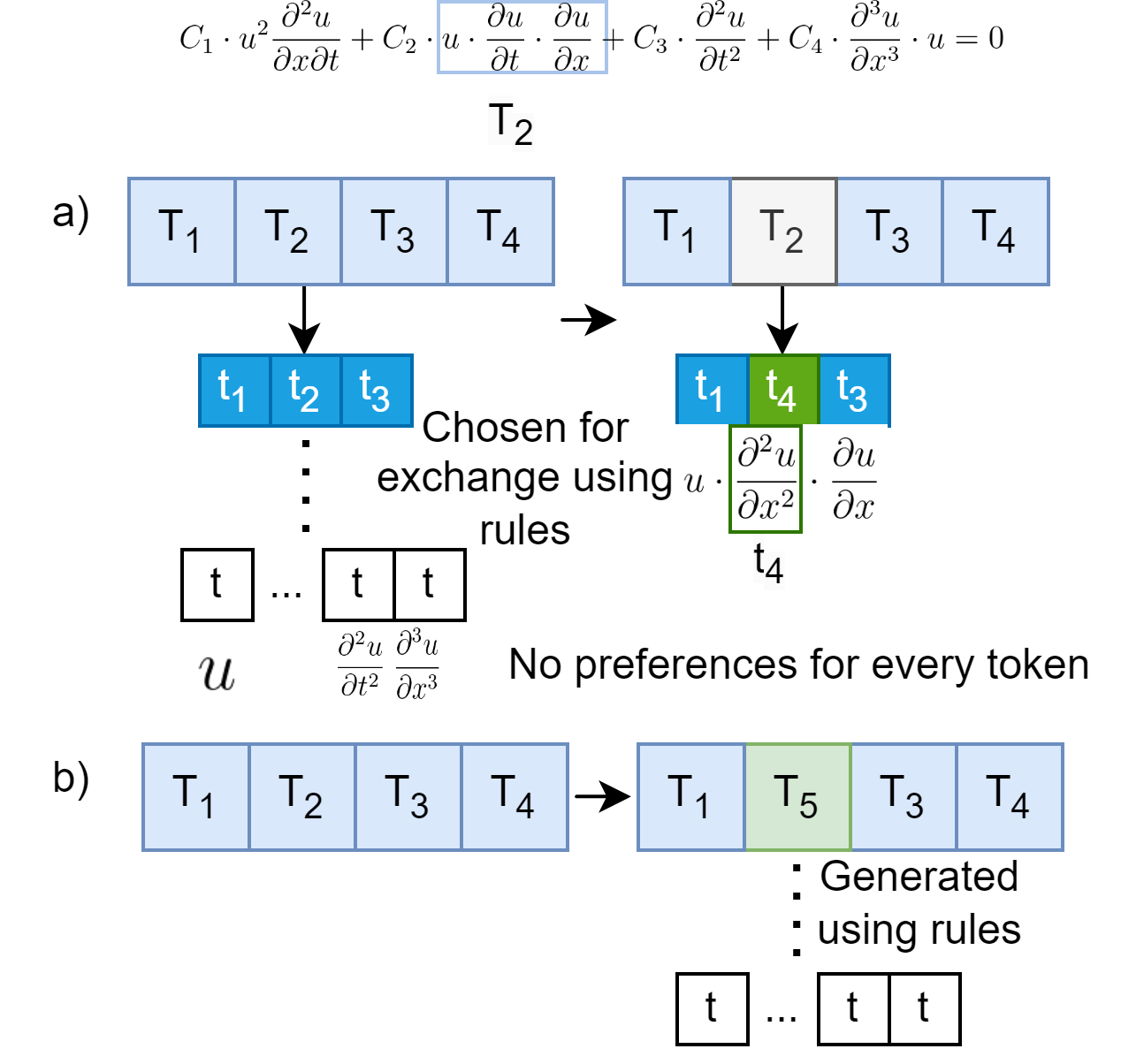}
    \caption{The classical algorithm mutation. New tokens a) and new term b) are generated using a uniform distribution.}
    \label{fig:uniform_mutation}
\end{figure}

The mutation operator has two forms -- term exchange and token exchange -- that could be applied with a given pre-defined probability, as shown in Fig.~\ref{fig:uniform_mutation}.

The token exchange shown in Fig.~\ref{fig:uniform_mutation}a) is simply replacing one token with another using the homogeneous pool of tokens. For the term exchange (Fig.~\ref{fig:uniform_mutation}b), the new term is generated using the homogeneous pool of tokens: first, the length of the token is chosen randomly, and second, the tokens are chosen from the pool. 

In summary, as input set of observations $U$ in the grid $X$ and set of token families $\Phi$ are used. As a result, we get an expression in the form of differential operator $Lu=f$. The result could be ODE, PDE, and their system depending on a dimensionality of $U$ and $X$.

The classical algorithm is realized as an evolutionary optimization framework \cite{maslyaev2021multi,maslyaev2022solver}, and the following modifications are also performed as part of a framework modification.

\section{Directed evolutionary search. Usage and extraction of background knowledge.}
\label{sec:modified_alg}

In this section, we finalize the algorithm described first in \cite{ivanchik2023directed}. It is an evolutionary differential equation discovery algorithm based on both EPDE and directed evolutionary serach. The algorithm is briefly outlined in Section.~\ref{sec:modified_operators}.

The main actor in the algorithm is the term preference distribution. It is the form in which background knowledge is expressed. Instead of parametric fixed equation form, we set the term preference that is likely to appear in equation, retaining the possibility to obtain all possible equations. Such preference distribution may be chosen expertly and automatically.

In Section~\ref{sec:proba_generation} we describe the automated term preference distribution extraction process. It consists of initial guess generation and following distribution forming algorithm.

\subsection{Modified evolutionary operators}
\label{sec:modified_operators}

The distribution of ``preferred'' tokens obtained automatically or imposed manually is then used to generate new tokens for the model in mutation and cross-over operators.

\begin{figure}[ht!]
    \centering
    \includegraphics[width=0.85\linewidth]{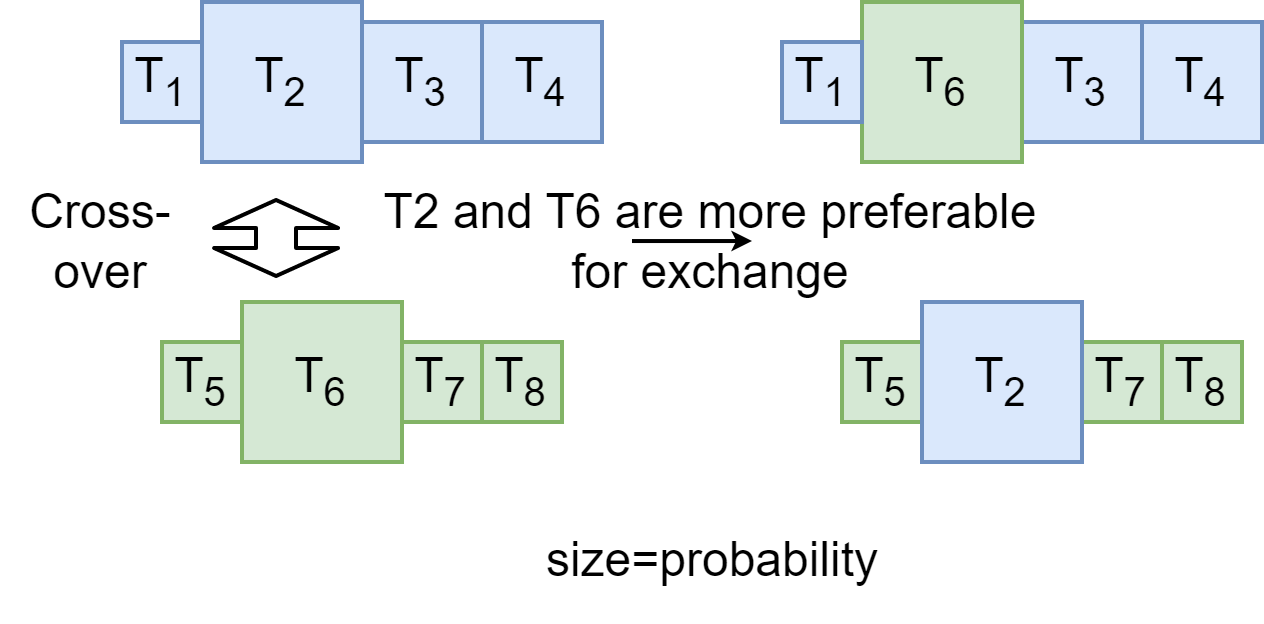}
    \caption{Modified cross-over. Terms have a different probability of participating in the cross-over; for illustration, the most probable terms win.}
    \label{fig:modified_cross_over}
\end{figure}

The modified cross-over operator uses two model structures to generate new ones, for every model the distribution of "preferred" tokens is generated separately. After that, tokens of every model are sorted with respect to complement model preferences using the probabilities as weights. Terms with higher weight have a higher chance of participating in cross-over exchange, as shown in Fig.~\ref{fig:modified_cross_over}.

\begin{figure}[ht!]
    \centering
    \includegraphics[width=0.85\linewidth]{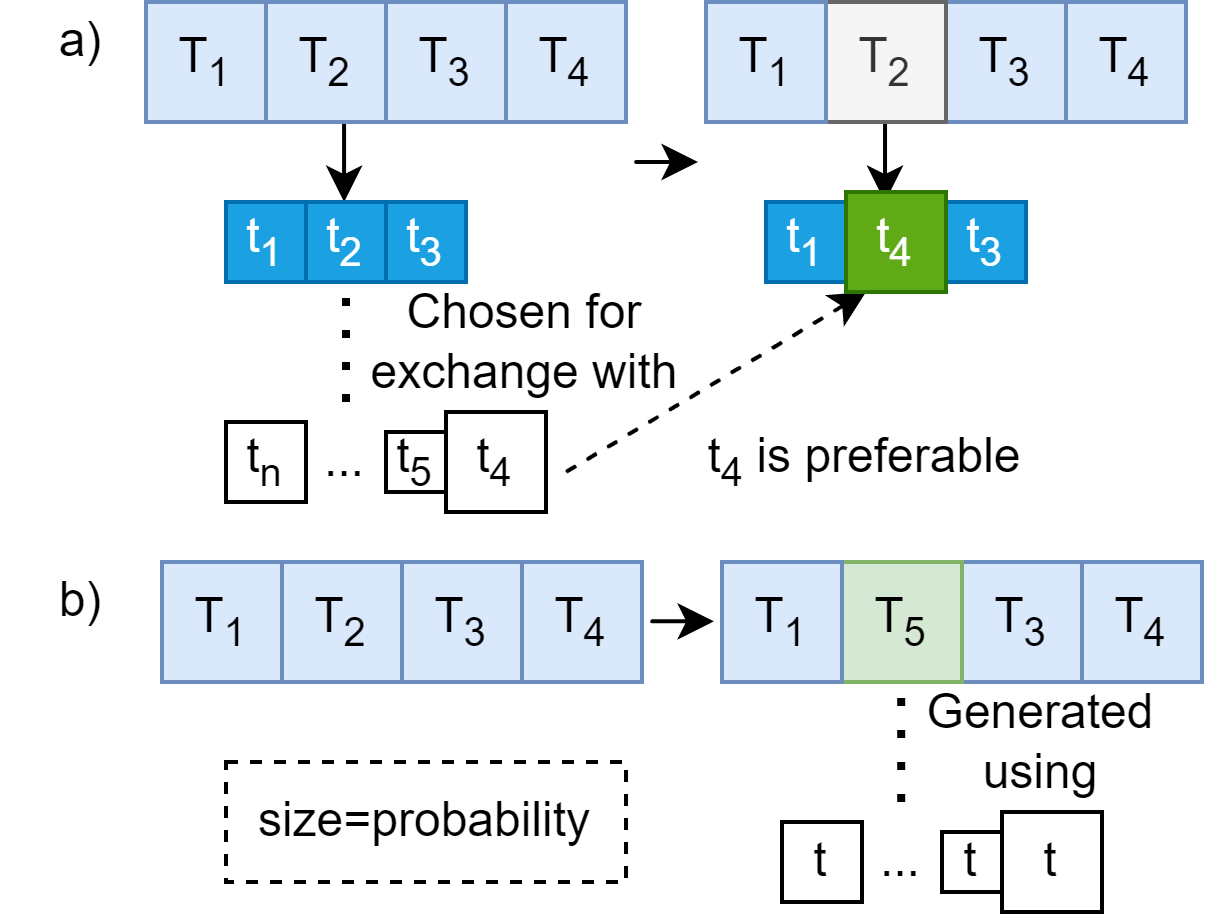}
    \caption{Modified mutation. New token a) is chosen using the importance distribution (for illustration, the most probable token is taken), and new term b) is generated using the importance distribution.}
    \label{fig:modified_mutation}
\end{figure}

The mutation operator takes all terms in the model structure as subject to mutation uniformly, but for a replacement token choice (Fig.\ref{fig:modified_mutation} a)), its generated importance distribution is taken into account. Furthermore, for term mutation, a new term is generated using the importance distribution as shown in Fig.~\ref{fig:modified_mutation} b).

\begin{figure}[ht!]
    \centering
    \includegraphics[width=0.8\linewidth]{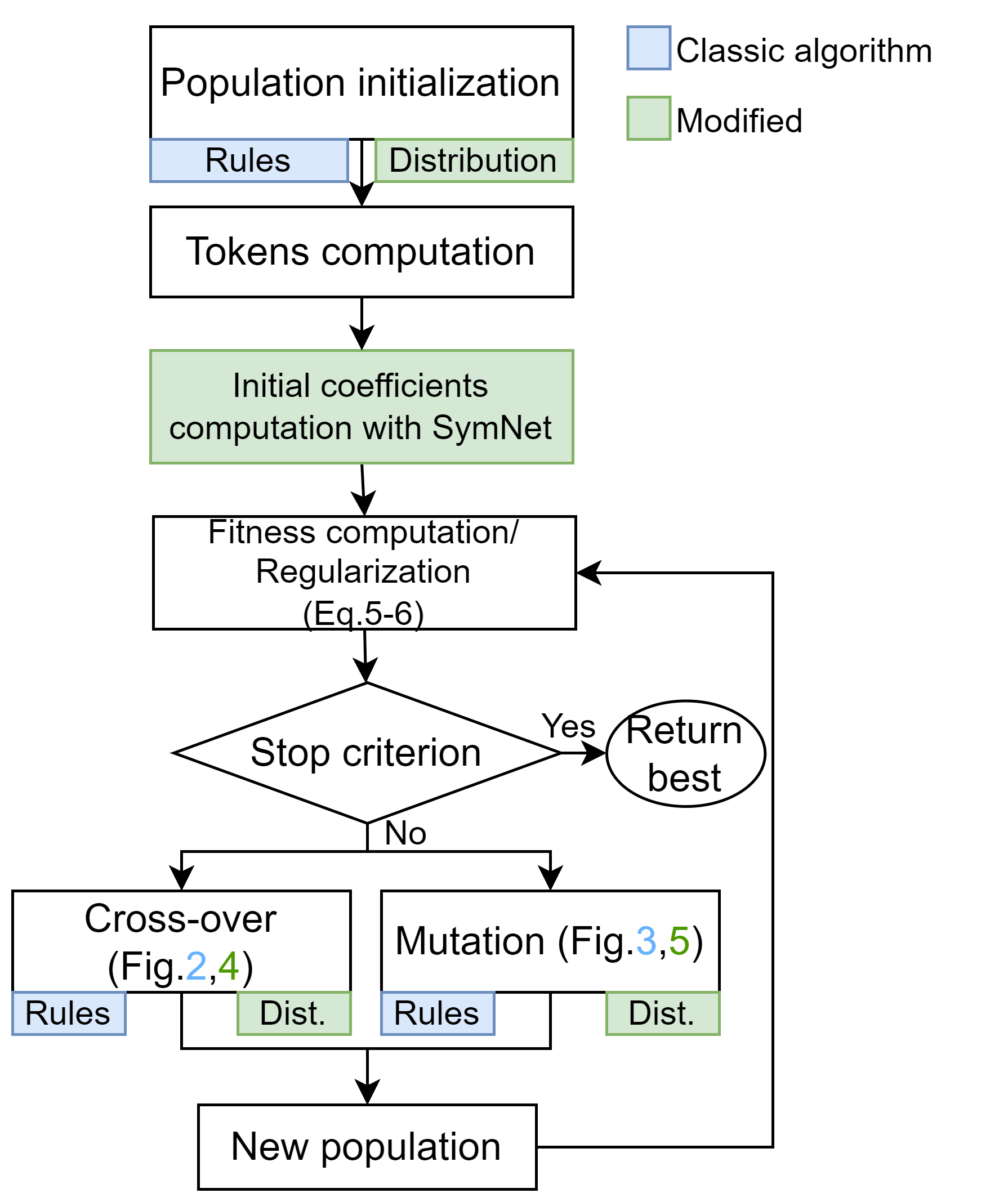}
    \caption{General scheme of classical and modified algorithms.}
    \label{fig:algorithm_scheme}
\end{figure}

In summary, the classical and modified algorithms have nearly the same structure as shown in Fig.~\ref{fig:algorithm_scheme}. The most substantial difference being that the modified algorithm has several extra steps for terms importance distribution calculation.  

We note again that the main changes are done to avoid the rigid restrictions on a structure optimization space $\Sigma'$. Instead, the applied expert background knowledge is used to change the geometry of the space $\Sigma'$, which in the case described in the paper is defined by the terms probability measure.

Proposed approach may be considered as the change in the geometry of the space $\Sigma'$ itself. Since the evolutionary algorithm is stochastic, changing the probability measure within the space is natural. The probability space $(\Sigma, \Sigma', p)$ introduced in a classical EPDE algorithm has a conventionally (rules make it shifted) uniform measure $p$. However, using a priori knowledge, we may introduce the 'importance' factor of the term to be added to the model.

\subsection{Probability distribution generation} 
\label{sec:proba_generation}

We used a modified SymNet architecture as an educated initial guess of the structure of the equation. We describe the modifications in \ref{app:modified_symnet}. We note that the initial guess may be obtained by various means, the described approach works with an arbitrary initial guess in form of the differential equation.  { There are several discussion points on this choice. The first is if the bad initial guess makes the end result worse. Second, is there are any other choices available.

There are overall two incomes of bad choice -- initial guess that contains wrong parts and initial guess that contains significant number of the possible terms within the guessing algorithm. As part of the answer to the first question, in the following, we show that SymNet does not always give a good guess (see \ref{app:symnet_results}). However, the quality of the search (robustness and number of iterations) was still improved. Second, if the guessing algorithm is not able to find any structure, the answer tends to be the overfitted equation that contains excessive number of terms. In this case, the uniform term distribution will be extracted, and thus the algorithm is not affected, i.e. works in a classical uni-direction mode.

Symnet was chosen since it is relatively fast and gives the initial guess and could generate a structure space $\Sigma'$ larger than that of SINDy and other regression methods, without an explicit definition of the terms. However, any equation discovery method could be taken instead, the described method in general does not depend on a guessing algorithm choice. We note that the quality of the initial guess undoubtedly affects the end result. However, as we show below, we could achieve quality increase when only part of the equation is guessed correctly.

}

The first step is done within the initial population generation stage. Firstly, all possible terms with a given EDPE hyper-parameters are generated. Secondly, if the term space of the initial guess does not coincide with the EPDE model term space, the mapping of the initial guess to the model term space is performed.

\begin{figure}[ht!]
    \centering
    \includegraphics[scale=0.2]{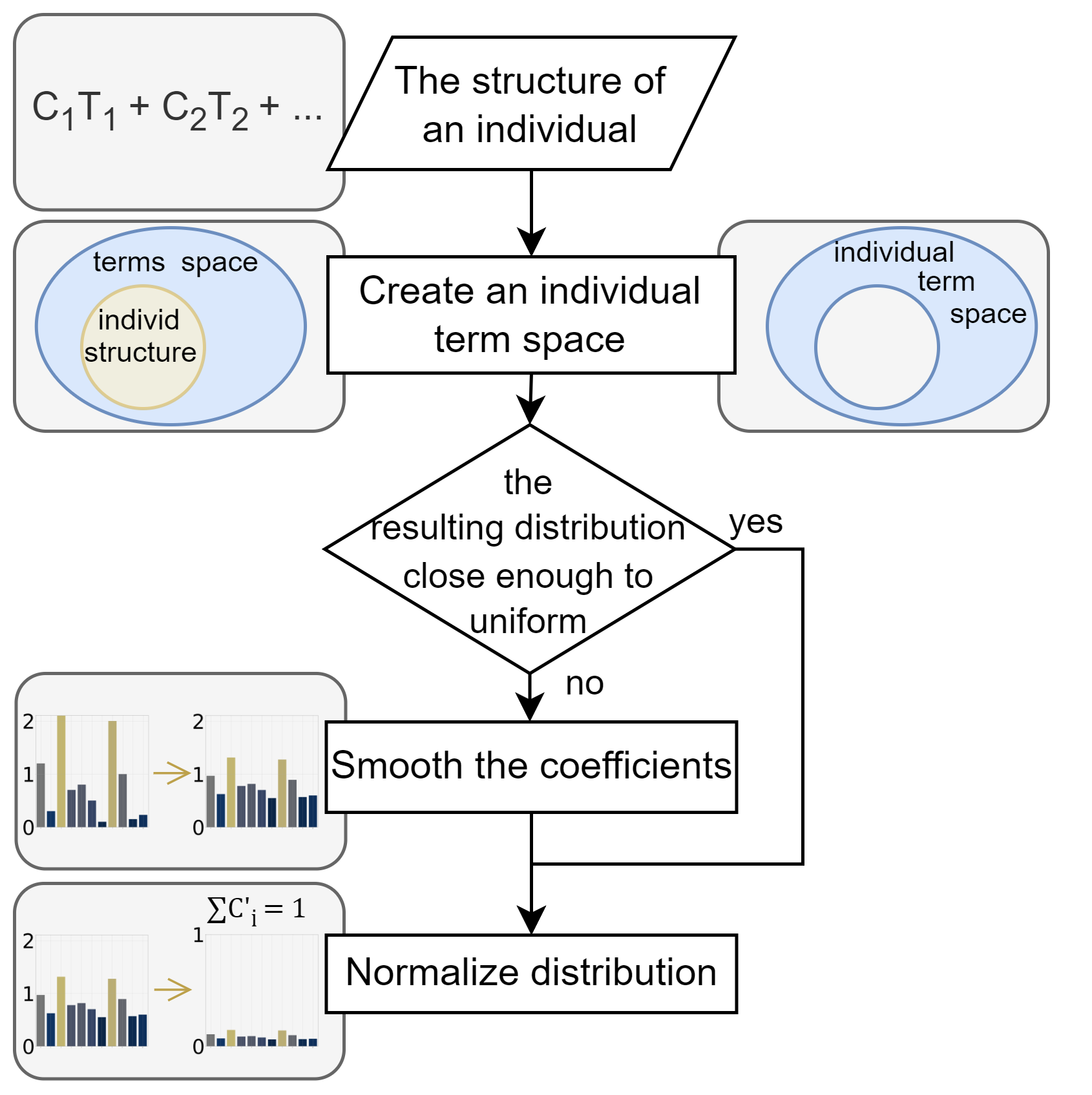}
    \caption{The scheme of the algorithm for probability distribution calculation.}
    \label{fig:distr_calc}
\end{figure}

Due to the rules that allow one to avoid equalities of type $0 \equiv 0$, every model has its own set of restricted terms. Therefore, the following steps are conducted separately within each individual's mutation and cross-over phases. The steps are illustrated in Fig.~\ref{fig:distr_calc}. First, depending on the structure of an individual undergoing a mutation process, the term space is adjusted so that the structural elements of an individual are excluded from the space. In this manner, every individual acquires its own term space and, consequently, a term importance distribution.

Once the individual term space is defined, the relation of a maximum value of term coefficients to the coefficients minimum is checked. If it exceeds a specific value, the resulting distribution might have low probabilities for terms with coefficients close to a minimum value. Then, the evolutionary algorithm may take a long time to converge due to the lack of variance in the population. One possible solution for the problem mentioned above is to smooth the coefficients' values so that they are closer to the average of all coefficients. This way, the probability distribution computed after coefficients normalization will be more similar to the uniform distribution.

The algorithm~\ref{alg:prob_distr_calc} describes the smoothing procedure in detail. The main idea is to mix the coefficients with the vector, consisting of coefficients average, where the vector size is equal to that of the coefficients. In this manner, the new relation of maximum value to minimum is equal to 2.4 when the original coefficients relation is bigger than it; otherwise, the smoothing step is skipped, and the normalization based on coefficients weight is conducted. 

In the previous work \cite{ivanchik2023directed}, it was discovered that if the relation of coefficients is too large, the advantages of guided optimization disappear. To prevent this problem, the ratio of maximum to minimum is limited by a constant, which will be referred to as a mixing factor (mf). Another reason for introducing this parameter is that the initial guess can rarely coincide with the actual equation (refer to Sec.~\ref{sec:symnet_standalone}) and thus can only provide a guess about the equation. Therefore the mixing factor may also be interpreted as a measure of 'trust' that we assign to the initial guess.

The value of a mixing factor can be any number in [1.0, 5.0] with default value of 2.4. It reflects the ability of the meta-algorithm to discover the initial guess and can be corrected in accordance with its performance - the poorer it performs the lower the mixing factor should be. The upper bound is set to be 5.0 due to the reason that higher relative differences in coefficients may obstruct the evolutionary optimization process as it can be seen in \cite{ivanchik2023directed}.

\begin{algorithm}[ht!]
\label{alg:prob_distr_calc}
\KwData{Set of allowable terms - $terms\_allowed$; their initial coefficients - $coefs$; mixing factor mf (default: 2.4) }
\KwResult{probability distribution}
$coefs = abs(coefs)$\;
\eIf { $max(coefs) /  min(coefs) > mf$}{

            $min\_max\_f = \text{mf} \cdot min(coefs) - max(coefs)$\;
            $smooth\_f =min\_max\_f / (\min\_max\_f - (\text{mf} - 1) \cdot average(coefs))$\;
            $vec\_average = average(coefs)$\;
            $smoothed\_coefs = (1 - smooth\_f) \cdot coefs + smooth\_f \cdot vec\_average$\;
            $probabilities = smoothed\_coefs / \sum {smoothed\_coefs}$\;
            
        }{
            $probabilities = coefs / \sum {coefs}$\;
        }
 \caption{The pseudo-code of probability distribution calculation}
\end{algorithm}

\begin{figure}[ht!]
    \centering
    \includegraphics[scale=0.3]{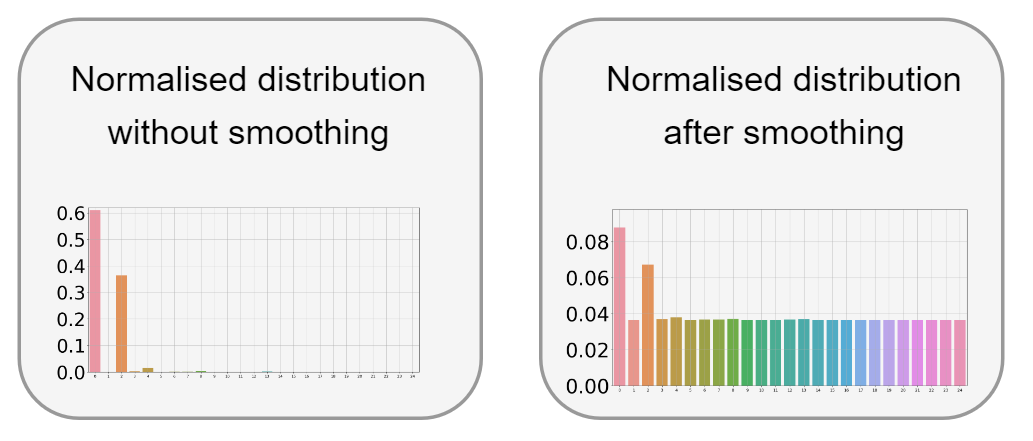}
    \caption{Example of distributions with and without smoothing procedure.}
    \label{fig:smooth_importance}
\end{figure}

The importance of the smoothing procedure is explained in Fig.~\ref{fig:smooth_importance}. The example is given for a set of coefficients with 25 values, where the maximum is equal to 0.52 and the minimum is $4 \cdot 10^{-7}$. In Fig.~\ref{fig:smooth_importance}, the left distribution results from the direct normalization of the coefficients, while the distribution on the right is obtained after applying the smoothing step. From the graphs, it can be concluded that the smoothing procedure reduces large coefficients and increases small values, where the processes of reduction and increasing are limited by coefficients average as the bottom and the upper bound correspondingly. In addition, the relative relations between the coefficients are preserved. Hence, in the provided example, columns 0 and 2 still have the largest values, while 2 is smaller than 0; simultaneously, the fourth and eighth columns are slightly larger than the others.

\begin{figure}[ht!]
    \centering
    \includegraphics[scale=0.3]{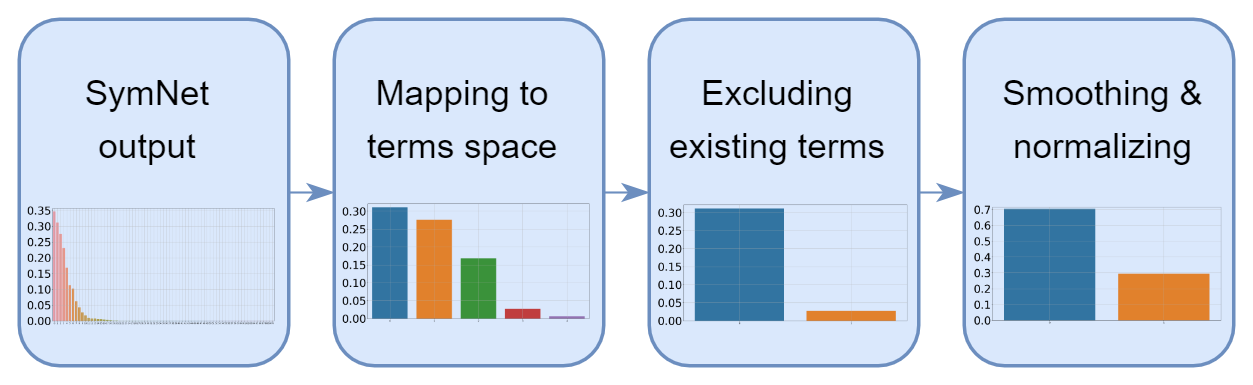}
    \caption{Example of distribution calculation for one individual, wave equation.}
    \label{fig:individ_distr}
\end{figure}

An example of calculating the probability distribution of an individual in the problem of finding wave equations is given in Fig.~\ref{fig:individ_distr}. The initial guess contains 71 non-zero terms with the precision of 16 digits after the decimal point, while the evolutionary algorithm terms space only has 5. Therefore, the mapping of terms is conducted. Given a term space, an individual already had three out of five terms during the mutation process. In order to obtain a new, unique term, the term space is reduced. The maximum value of coefficients after reducing the terms space is 0.3; at the same time, the minimum is only 0.021; because their relation is too significant, the smoothing procedure is applied. The values are normalized to ensure the probability distribution property. Namely, the sum of all probabilities must be equal to 1.

\section{Experimental results}
\label{sec:experiments}

The goal of the experiments is to determine whether the knowledge guidance mechanism in the form of the initial guess of term distribution can lead to better optimization in terms of stability, accuracy, and speed. The experiments are conducted for three classes of equations: Burgers', wave, and Korteweg--de Vries. The data for experiments was obtained either analytically or numerically, the initial-boundary problem statements and solution methods are placed in \ref{app:IVP_statements}.

If possible, the results are compared with that of the PySINDY \cite{desilva2020} framework (ver. 1.7.5); if not, only two algorithms are considered: classical and modified.

\subsection{Experimental setup}
\label{sec:experimental_setup}

For every experiment, we run the classical and modified algorithms fifty times. We use the PySINDY package on the same data.

The performance of the algorithms is measured with different noise levels in the data. Equation~\ref{eq:noising} describes adding noise of a certain magnitude to the data. Magnitudes differ in scale depending on the input data. For this reason, every type of equation has its limit magnitude, under which the classical algorithm cannot discover the desired equation in any of fifty runs. Namely, as stated above, the observational data have the form $U=\{u^{(i)}=(u^{(i)}_1,...,u^{(i)}_L)\}_{i=1}^{N}$. The noise levels are relative to the limit case and therefore are equal to 0\%, 25\%, 50\%, 75\% and 100\% of the limit noise level.

\begin{equation}
\begin{array}{cc}
     u_{noised}^{(i)} = u^{(i)} + \epsilon^L, \;\; \epsilon \sim N(0, magnitude \cdot |u^{(i)}_j|)   \\
     \epsilon^L=\epsilon \times \epsilon \times ... \times \epsilon 
\end{array}
    \label{eq:noising}
\end{equation}

The quality metrics used to measure algorithm performance are mean absolute error (MAE) between coefficients of the obtained equations and 'ground truth' coefficients of the theoretical model and algorithms' convergence time. If several solutions are available, the structure of the equations is first checked. If the structure is correct, then MAE is computed. Among the computed MAEs, the minimum is selected as the final metric of the run.

Initially, with the double precision, there were around 100 different MAE values; after grouping them with the precision of 3-5 digits past the decimal point, we are left with approximately 3-25 clusters, each represented with some shade of green on the plotted figures. Clusters are placed in ascending order on the vertical axis by the MAE value around which they are clustered.

In addition, hyper-parameters for all experiments are presented in supplementary material in Tab. ~\ref{tab:hyperepde} and ~\ref{tab:hyperpysindy}.

\subsection{Initial guess: SymNet performance}

{
\label{sec:symnet_standalone}

In the proposed approach the SymNet architecture is used in order to provide an educated guess to EPDE. Although one might be tempted to use SymNet directly to discover the equations, in reality, some difficulties might be faced while doing so. This subsection is dedicated to these challenges and explains the reasoning behind SymNet' role in overall approach. The detailed equation and analysis placed in \ref{app:symnet_results}. Integral results are places in Tab.~\ref{tab:symnet_mae} for MAE (coefficient error).

\begin{table}[!ht]
\centering
\caption{Statistical MAE, SymNet results for all experiments}
\label{tab:symnet_mae}
\begin{tabular}{|c|c|c|c|c|}
\hline
\begin{tabular}[c]{@{}c@{}}Burgers'\\ eq. with\\ viscosity\end{tabular} &
  \begin{tabular}[c]{@{}c@{}}Burgers'\\ equation\end{tabular} &
  \begin{tabular}[c]{@{}c@{}}Wave \\ equation\end{tabular} &
  \begin{tabular}[c]{@{}c@{}}KdV \\ equation\end{tabular} &
  \begin{tabular}[c]{@{}c@{}}Inhomogeneous \\ KdV \\ equation\end{tabular} \\ \hline
\begin{tabular}[c]{@{}c@{}}$0.0058$ \\ $\pm 0.0008$\end{tabular} &
  \begin{tabular}[c]{@{}c@{}}$0.0239$ \\ $\pm 0.0194$\end{tabular} &
  \begin{tabular}[c]{@{}c@{}}$0.0435$ \\ $\pm 0.0545$\end{tabular} &
  \begin{tabular}[c]{@{}c@{}}$0.1118$ \\ $\pm 0.0001$\end{tabular} &
  \begin{tabular}[c]{@{}c@{}}$0.1497$ \\ $\pm 0.0214$\end{tabular} \\ \hline
\end{tabular}
\end{table}

In Tab.~\ref{tab:symnet_shd} for structural Hamming distance (SHD) results are shown, i.e. how many terms we should add or remove to reach ground truth equations.

\begin{table}[!ht]
\centering
\caption{Statistical SHD, SymNet results for all experiments (with the precision of 6 digits)}
\label{tab:symnet_shd}
\begin{tabular}{|c|c|c|c|c|}
\hline
\begin{tabular}[c]{@{}c@{}}Burgers'\\ eq. with\\ viscosity\end{tabular} &
  \begin{tabular}[c]{@{}c@{}}Burgers'\\ equation\end{tabular} &
  \begin{tabular}[c]{@{}c@{}}Wave\\ equation\end{tabular} &
  \begin{tabular}[c]{@{}c@{}}KdV\\ equation\end{tabular} &
  \begin{tabular}[c]{@{}c@{}}Inhomogeneous\\ KdV\\ equation\end{tabular} \\ \hline
$26 \pm 4$ &
  $19 \pm 4$ &
  $52 \pm 9$ &
  $29 \pm 5$ &
  $103 \pm 30$ \\ \hline
\end{tabular}
\end{table}

Overall, the SymNet algorithm provided adequate results in cases of simple equations, but was inaccurate for more complex cases due to low coefficient error, but high SHD.

The experiment with wave equation deserves particular attention. We note that if the balancing term is represented by the second space derivative, the equation is obtained without fail. This leads us to the conclusion that if in the ground truth equation the coefficients have small absolute values, the accuracy of the algorithm might not be high.

Considering all of the above, SymNet might be used as a starting point for another equation discovery algorithm, and although it is able to provide the solution by itself, its quality may be severely lacking.

}

\subsection{Parameter sensitivity: mixing factor}
\label{sec:distr_sensitivity}

{ In order to capture the dependency of the modified algorithm efficiency on the changes in term importance distribution, experiments with different mixing factors are performed. The results of these experiments are compared with the results of the algorithm with optimal distribution. Optimal distribution referrers to a distribution that is the most close to the ideal one, where the importance of the desired terms is 3.0 to 3.5 times higher than the other terms. Optimal distribution $Q(\text{mf})$ is obtained by SymNet and then processed with a mixing factor, acquired by Eq.~\ref{eq:optimal_distr}.}

\begin{equation}
    \text{mf}_{opt} = \operatorname*{arg\,min}_{\text{mf}} |D_{KL}(P||Q(\text{mf}))|
    \label{eq:optimal_distr}
\end{equation}

{

From the MAE results depicted in the following sections we can conclude that depending on the value of a mixing factor, the modified algorithm may yield different results. Therefore, the purpose of this section is to study this dependency.

The modified algorithm was run with several mixing factors - 3.0, 3.6, 4.5 and 2.4 by default. The mixing factors defined by Kullback-Leibler generally slightly differ by the noise level in the data. %In the tables ~\ref{tab:mix_kdv_s}-~\ref{tab:mix_wave} the numbers of successful runs are presented. 
In \ref{app:mixing_factor_tables} detailed tables for mixing coefficient sensitivity analysis are provided.
With the data on considered types of equations we can conclude, that the tuning of mixing factor may provide extra benefits, specifically, the mixing factors derived from proximity of ideal and found distributions, are mainly the most optimal.

In all experiments below, we use two different mixing factors when possible -- 2.4 by default and the optimal one.

}

\subsection{SINDy comparison}

%\subsection{Comparison with PySINDy: Burgers' equation}
%\label{sec:burg_result}

We ran a classical algorithm for experiments and modified it with the setup described in Sec.~\ref{sec:modified_alg}. For these experiments PySINDy framework was used as well.

\begin{figure}[ht!]
    \centering
    \includegraphics[width=0.9\linewidth]{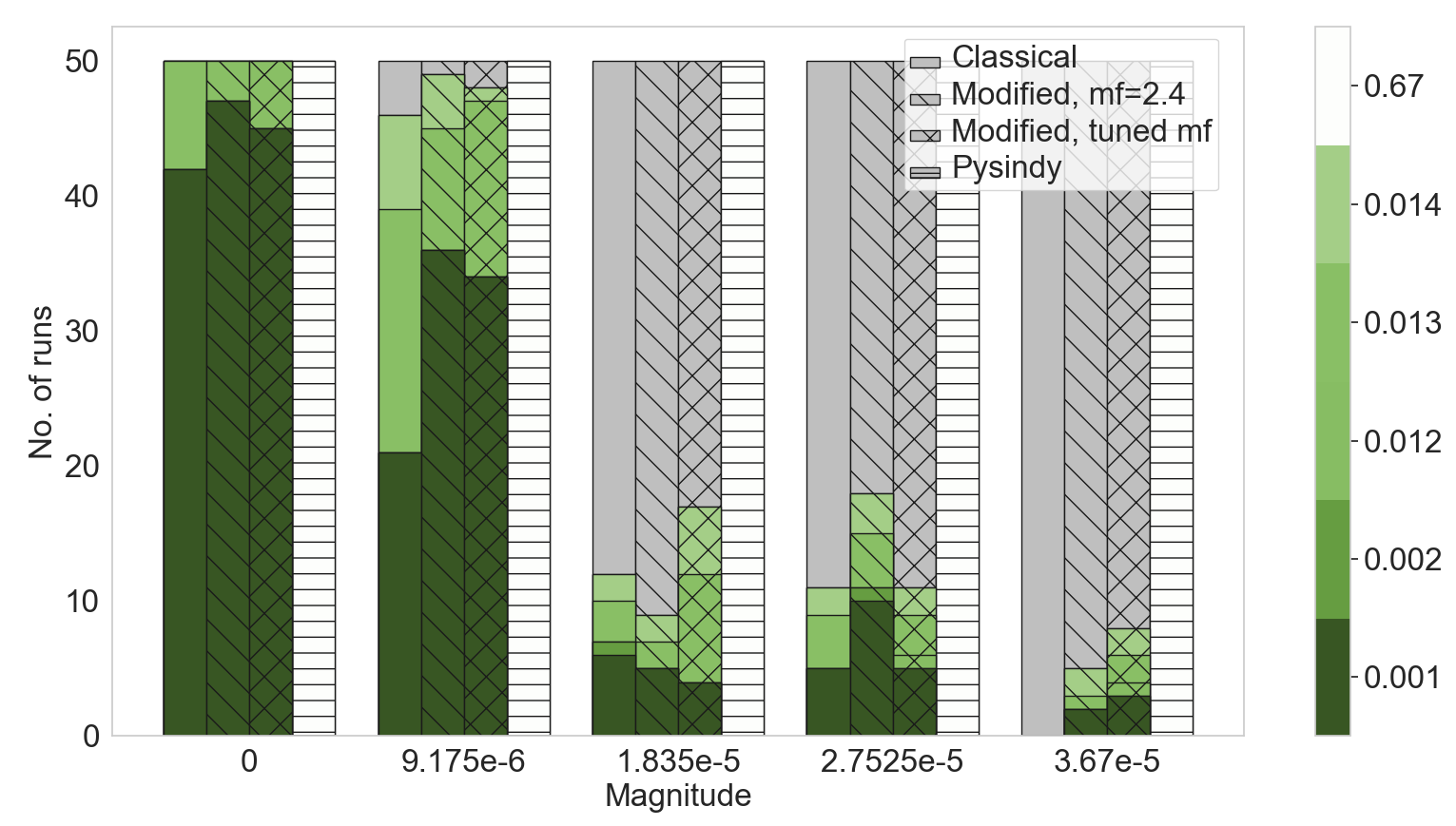}
    \caption{Mean absolute error distributions for fifty runs for different noise magnitude values, inviscid Burgers' equation. The columns with no hatch represent the classical algorithm, "$\backslash$" hatch - modified with mf equal to 2.4, "x" hatch - modified with tuned mf, "--" hatch - PySINDY. Runs where the equation was not obtained are depicted in gray.}
    \label{fig:burgers_mae}
\end{figure}

In Fig.~\ref{fig:burgers_mae} - inviscid Burgers equation case, it can be seen that the bars of the modified algorithm tend to have more runs with darker green colors. We can conclude that the modified algorithm performs better than the classical one regarding precision. Considering the number of runs with gray outcomes, we can conclude that the modified algorithm is comparatively more stable.

{ We note that although PySINDy framework was able to obtain a correct equation, the accuracy of the algorithm was relatively low considering the complexity of the Burger's equation.}

\begin{figure}[ht!]
    \centering
    \includegraphics[width=0.9\linewidth]{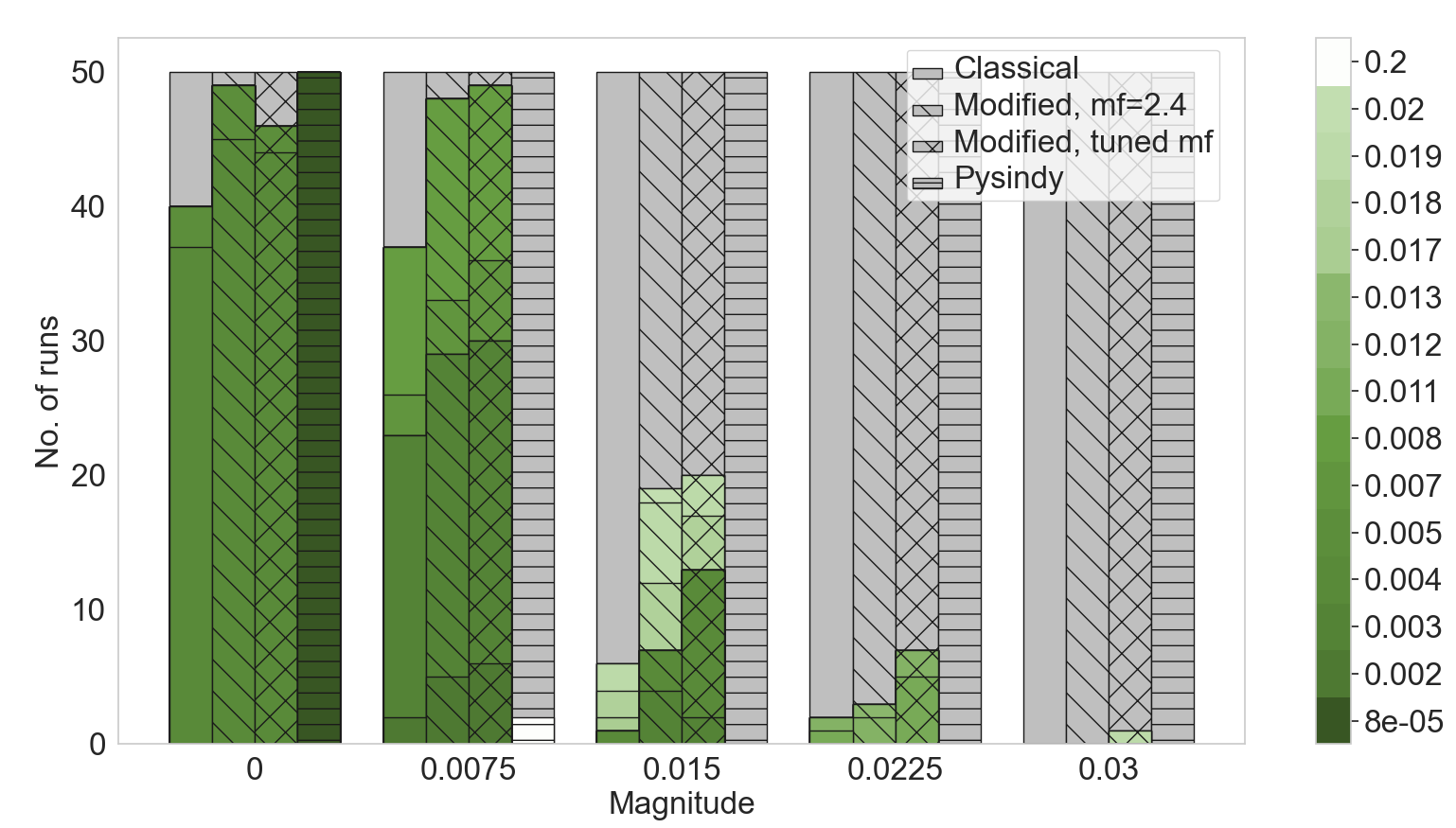}
    \caption{Mean absolute error distributions for fifty runs for different noise magnitude values, Burgers' equation with viscosity. The columns with no hatch represent the classical algorithm, "$\backslash$" hatch - modified with mf equal to 2.4, "x" hatch - modified with tuned mf, "--" hatch - PySINDY. Runs where the equation was not obtained are depicted in gray.}
    \label{fig:burgers_s_mae}
\end{figure}

 The experimental results of the MAE measured coefficients in case of Burgers' equation with viscosity term are shown in Fig.~\ref{fig:burgers_s_mae}.{ Comparing the classical algorithm to the modified one, we can surmise that the modified algorithm outperforms the classical one in terms of stability and accuracy. The tuning of mixing factor further increases the stability of the modified algorithm with the exception of no noise level. PySINDy algorithm shows excellent stability and accuracy in the case of zero noise magnitude but drastically loses its advantages once the noise magnitude is raised. }

\begin{figure}[ht!]
    \centering
    \includegraphics[width=0.9\linewidth]{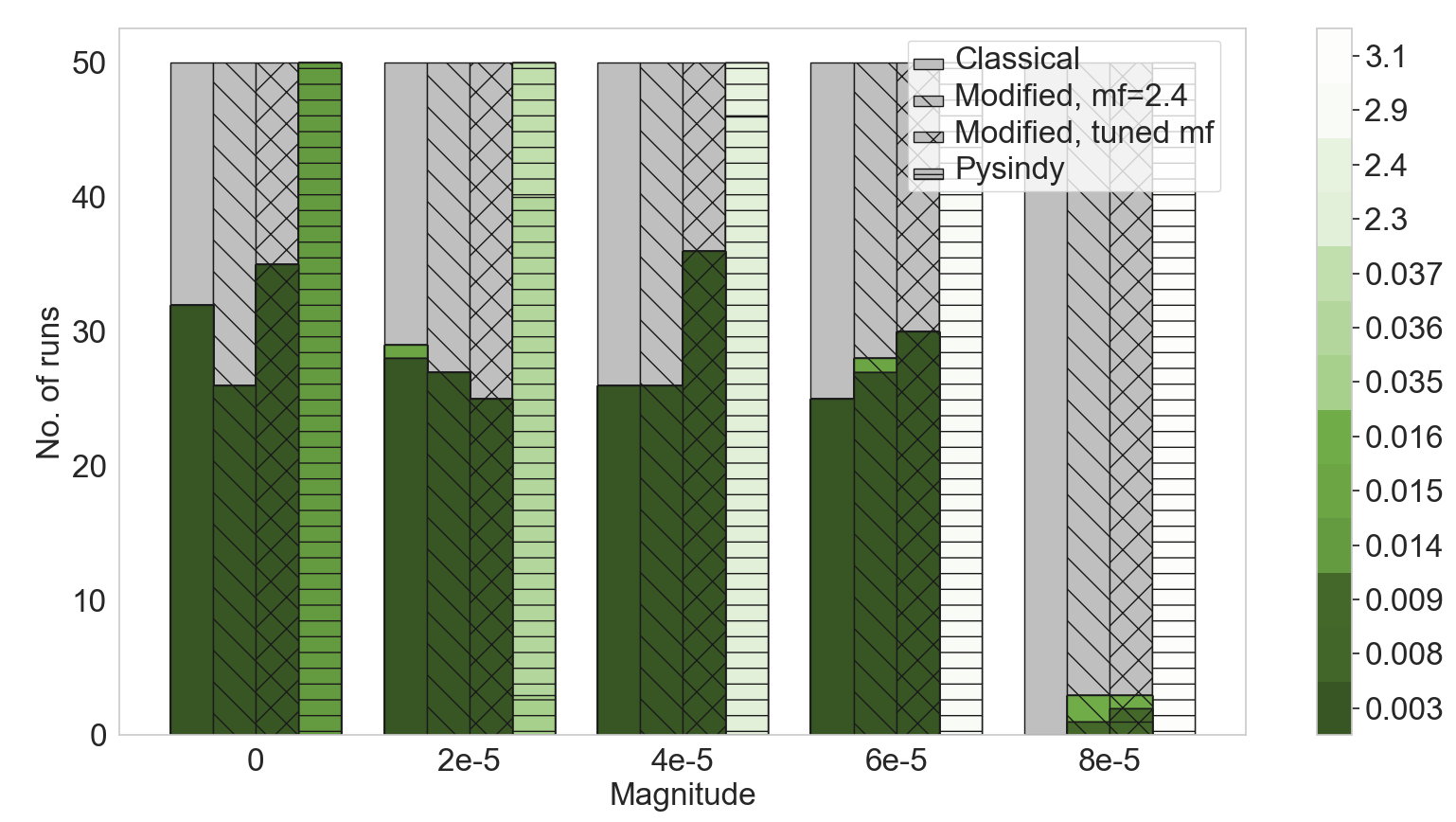}
    \caption{Mean absolute error distributions for fifty runs for different noise magnitude values, Korteweg -- de Vries equation. The columns with no hatch represent the classical algorithm, "$\backslash$" hatch - modified with mf equal to 2.4, "x" hatch - modified with tuned mf, "--" hatch - PySINDY. Runs where the equation was not obtained are depicted in gray.}
    \label{fig:kdv_s_mae}
\end{figure}

The homogeneous Korteweg -- de Vries equation was chosen as a more challenging case. The results for algorithms' MAE in coefficients are illustrated in Fig.~\ref{fig:kdv_s_mae}. { Unlike in viscous Burgers' equation, we can observe a decrease in the modified algorithm's stability, however, the tuning procedure could aid in its improvement.} The modified algorithm might be slightly better in extreme cases than classical one. In addition, the PySINDY algorithm remains robust with the increasing noise magnitude. The precision of both the classical and modified algorithms is relatively similar, whereas the PySINDY algorithm abruptly loses accuracy with increase in noise.

{ In conclusion, the modified algorithm often presents better stability and accuracy of the results, especially, if tuning procedure is carried out.} However, the time consumption of the approach is generally greater. The stability of PySINDY in comparison with the mentioned algorithms is still being determined; more experiments are needed to draw any conclusions, whereas the accuracy of the PySINDY package is mostly compromised when the noise magnitude rises.

\subsection{Additional tests}

\paragraph{Wave equation} Due to the hyperbolic nature of wave equations, it is rarely used to test equation discovery frameworks. However, the proposed approach can find time derivatives of any order. We note that this type of equation cannot be obtained with PySINDy because the algorithm restricts the equation's form: $u_t=F(x,u_x,u_xx,...)$.

Nevertheless, the running time (Fig.~\ref{fig:wave_time}) of the modified algorithm is significantly higher, apparently due to the reason that the SymNet module is run several times to compare losses of equations with $u_t$ and $u_{tt}$ terms, that are used to balance the right side of the equation.

\begin{figure}[ht!]
    \centering
    \includegraphics[width=0.9\linewidth]{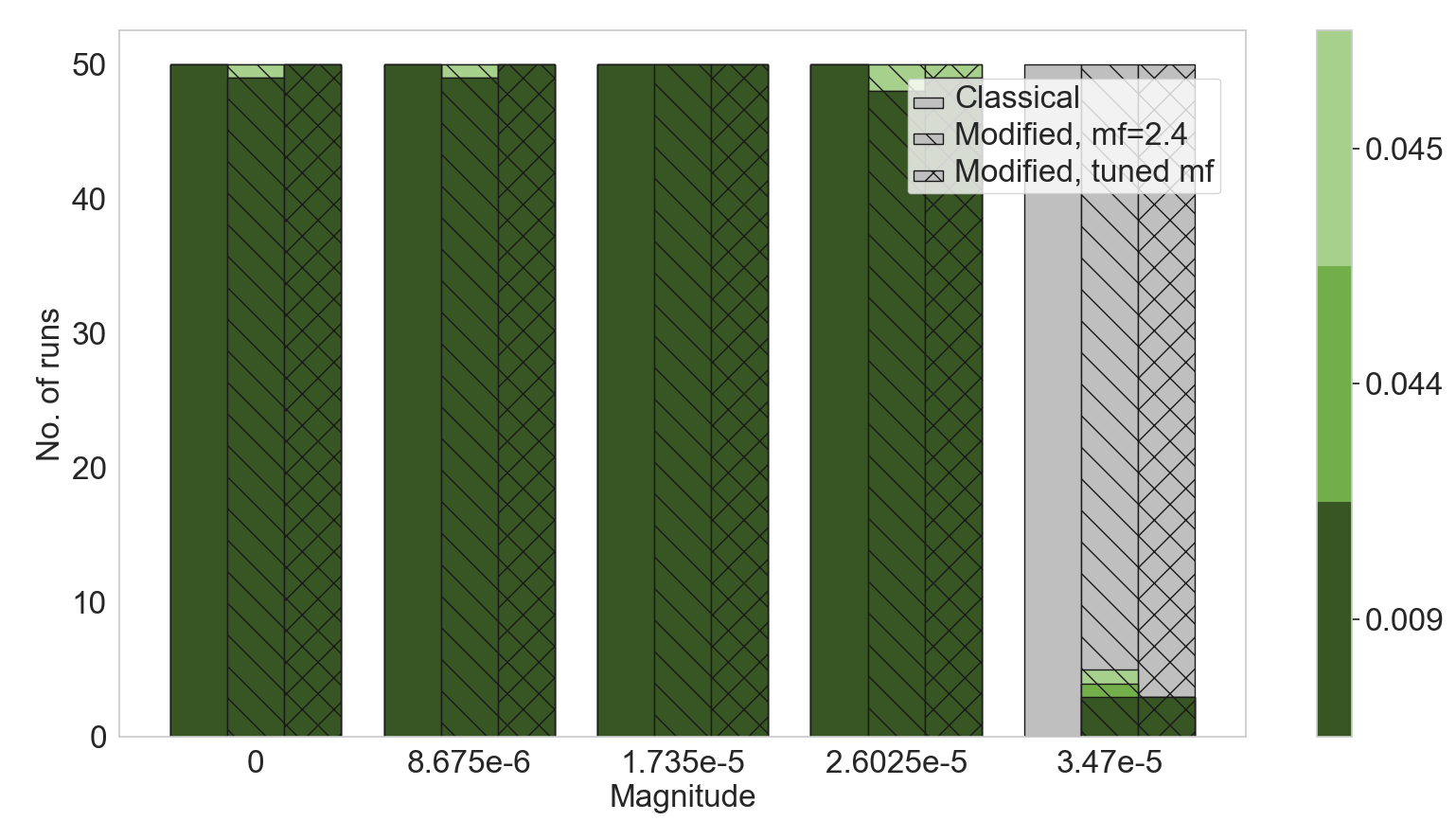}
    \caption{Mean absolute error distributions for fifty runs for different noise magnitude values, wave equation. The columns with no hatch represent the classical algorithm, "$\backslash$" hatch - modified with mf equal to 2.4, "x" hatch - modified with tuned mf. Runs where the equation was not obtained are depicted in gray.}
    \label{fig:wave_mae}
\end{figure}

The MAE distributions are shown on Fig.~\ref{fig:wave_mae}. Given the outcomes of the runs, it becomes clear that both classical and modified algorithms have similar coefficient error levels, although the modified algorithms might be slightly more stable. We note that the modified algorithm with tuned mixing factor does not have any significant difference in MAE with modified algorithm that has a mixing factor of 2.4.

\paragraph{Inhomogeneous Korteweg -- de Vries equation} The proposed approach was also tested on a more complex case in the equation discovery area - the inhomogenous Korteweg-de Vries equation. 

\begin{figure}[ht!]
    \centering
    \includegraphics[width=0.9\linewidth]{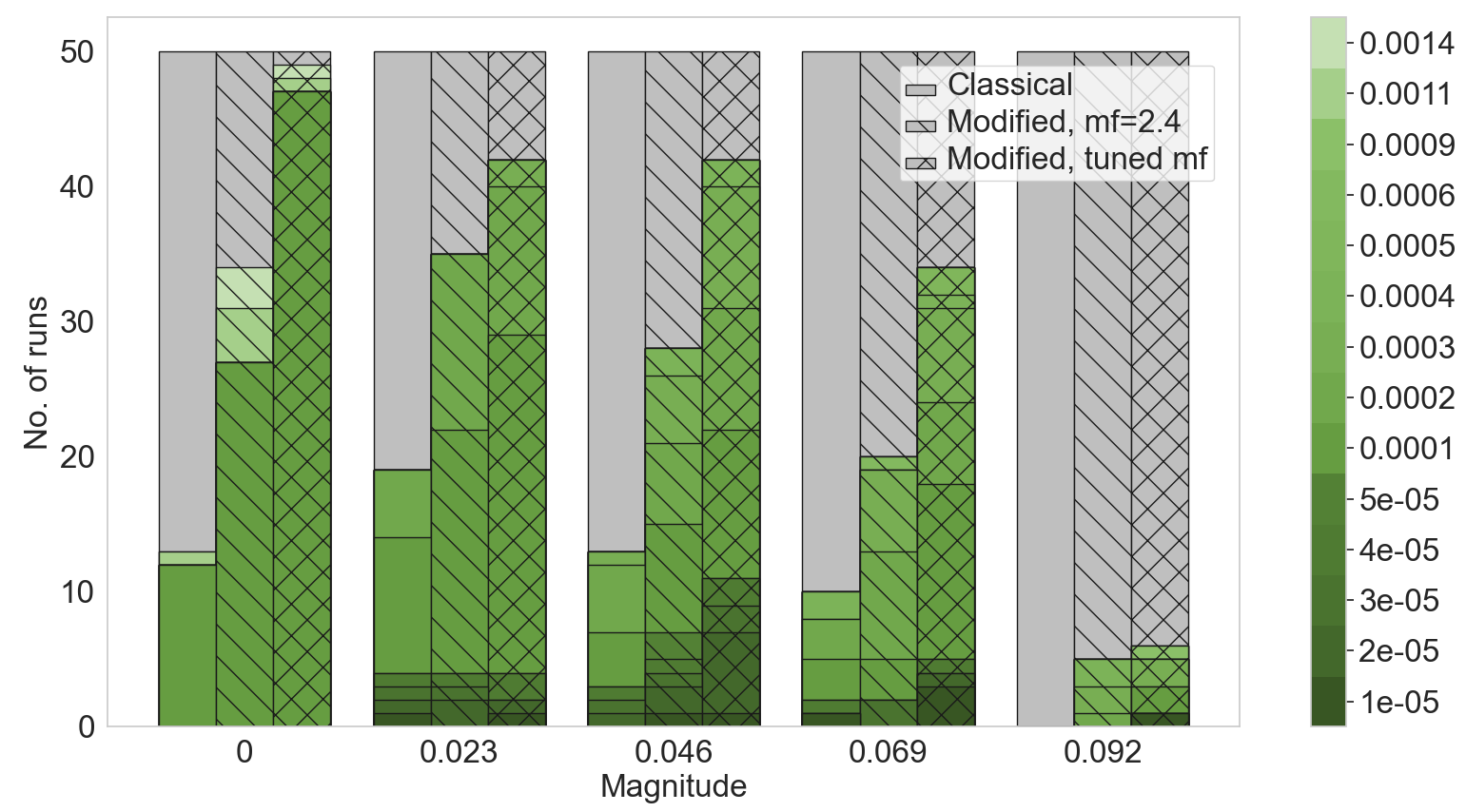}
    \caption{Mean absolute error distributions for fifty runs for different noise magnitude values, Korteweg -- de Vries equation. The columns with no hatch represent the classical algorithm, "$\backslash$" hatch - modified with mf equal to 2.4, "x" hatch - modified with tuned mf. Runs where the equation was not obtained are depicted in gray.}
    \label{fig:kdv_mae}
\end{figure}

In this experiment, two algorithms were compared: classical and modified. The results in the form of plots are shown in Fig.~\ref{fig:kdv_mae}. Unlike in the experiment with the homogeneous Korteweg -- de Vries equation, the modified algorithm was more precise, stable, and considerably faster than the classical one.

{ The benefits of tuning the mixing factor can also be seen on Fig.~\ref{fig:kdv_mae}. For the Korteweg -- de Vries equation, the tuning procedure yielded roughly 30\% increase in discovery rate averaged on all noise levels compared to the modified algorithm with a mixing factor of 2.4. This fact encourages us to further develop this idea, as currently it only serves the purpose of showing the potential of the modified algorithm.}

\subsection{Algorithms time consumption}

During all experiments execution time was recorded. Detailed reports may be found in \ref{app:time_consumption}.

To sum up -- the modified algorithm is generally slower than the classical one (reference time 5 s for classical EPDE and 15 s for modified algorithm). We note that the PySINDY algorithm's average running time is around 0.01 s, which outperforms both the classical and the modified algorithms. However, the equation structure's restrictions may outweigh its benefits.

\section{Discussion and conclusion}
\label{sec:conclusion}

The paper proposes a new way to introduce a priori applied area knowledge into the equation discovery algorithm without strict structure restrictions. It is done by introducing the geometry change in the model structure space by changing the probability measure used in the classical cross-over and mutation operators of the EPDE algorithm. We improve the algorithm's ability to converge towards a given equation more frequently. The distribution could be extracted in an automated manner using the proposed SymNet architecture training.

The main paper findings are:

\begin{itemize}
    \item Evolutionary algorithms could be used to incorporate knowledge more softly;
    \item Knowledge incorporation in the form of a term preference distribution leads to a more stable convergence;
\item Knowledge in the form of distribution could be extracted manually;
    \item Evolutionary algorithms work better for high noise magnitudes.
\end{itemize}

Compared to the classical gradient-based sparse regression approach, our proposed evolutionary approach offers more flexibility and produces restored equation quality of at least the same level. In addition, it can restore an equation that gradient algorithms might have found incorrectly in complex cases. However, one downside of the evolutionary approach is that the optimization time is typically longer. As the main advantage of gradient-based algorithms, one may name optimization speed.

Despite this, we observe that domain knowledge incorporation may increase the success rate of the algorithm by more than 40\% (Tab.~\ref{tab:incr_discrate_default}) in the case of an unbiased solution in the task of complex equation discovery (inhomogeneous Korteweg--de Vries equation); on the other hand, in the task of simple equations discovery there was no room for improvement, as the success rate reached the score of 1.0 in the classical algorithm. The noise introduction shows that the difference is more apparent for complex cases of differential equations. The proposed algorithm allows one to restore known equations with higher noise levels, increasing robustness on the average of 12.5\%, ranging from 2\% up to 32\%. 

{The proposed methodology may be further improved by tuning of mixing factor. In this case the increase in discovery rate (Tab.~\ref{tab:incr_discrate_kl}) in complex differential equation types may be enlarged to as far as 70\% for noise-free data and by 21.3\% on average, ranging from 2\% to 58\%, when the noise is added to the data.

The accuracy of all algorithms is presented in Tabs. ~\ref{tab:dfp_mae}, ~\ref{tab:dfs_mae}, ~\ref{tab:dfs_tuned_mae}, ~\ref{tab:dfo_mae}. As it may be noted the evolutionary approaches are generally more accurate than PySINDy framework. Whereas the accuracy of evolutionary algorithms is on par with one another with tuned modified algorithm having a slight advantage.

Nevertheless, we note that the modified algorithms do not always perform better than their classical form even when the tuning procedure is performed, which motivates us to conduct further research.}

\section{Code and data availability}

The code and data to reproduce the experimental results are openly available in the repository \url{https://github.com/ITMO-NSS-team/EPDE_automated_knowledge}

\section*{Acknowledgment}

This work was supported by the Analytical Center for the Government of the Russian Federation (IGK 000000D730321P5Q0002), agreement No. 70-2021-00141.

\bibliographystyle{elsarticle-num} 
\bibliography{references}

\newpage

\appendix

\section{Success rates of equation discovery}
\label{sec:ratings}

{The increase in success rate is calculated as follows: first, an average (on 50 runs) equation discovery rate is computed separately for each evolutionary algorithm and noise levels, after which the efficiency of the proposed methodology can be evaluated as the difference in success rate between the modified and classical algorithms.}

\begin{table}[]
\centering
\caption{Increase (\%) in equation discovery rate of modified algorithm compared to classical one, mixing parameters defined by KL}
\label{tab:incr_discrate_kl}
\begin{tabular}{|c|c|c|c|c|c|}
\hline
\begin{tabular}[c]{@{}c@{}}Noise \\ level\end{tabular} &
  \begin{tabular}[c]{@{}c@{}}Wave\\ equation\end{tabular} &
  \begin{tabular}[c]{@{}c@{}}Burgers'\\ equation\end{tabular} &
  \begin{tabular}[c]{@{}c@{}}Burgers'\\ equation\\ with\\ viscosity\end{tabular} &
  \begin{tabular}[c]{@{}c@{}}Homogeneous\\ KdV\\ equation\end{tabular} &
  \begin{tabular}[c]{@{}c@{}}Inhomogeneous\\ KdV\\ equation\end{tabular} \\ \hline
0\%   & 0 & 0  & 12 & 6  & 72 \\ \hline
25\%  & 0 & 4  & 24 & -8 & 46 \\ \hline
50\%  & 0 & 10 & 28 & 20 & 58 \\ \hline
75\%  & 0 & 0  & 10 & 10 & 48 \\ \hline
100\% & 6 & 16 & 2  & 6  & 12 \\ \hline
\end{tabular}
\end{table}

\begin{table}[]
\centering
\caption{Increase (\%) in equation discovery rate of modified algorithm compared to classical one, default mixing parameters}
\label{tab:incr_discrate_default}
\begin{tabular}{|c|c|c|c|c|c|}
\hline
\begin{tabular}[c]{@{}c@{}}Noise \\ level\end{tabular} &
  \begin{tabular}[c]{@{}c@{}}Wave\\ equation\end{tabular} &
  \begin{tabular}[c]{@{}c@{}}Burgers'\\ equation\end{tabular} &
  \begin{tabular}[c]{@{}c@{}}Burgers'\\ equation\\ with\\ viscosity\end{tabular} &
  \begin{tabular}[c]{@{}c@{}}Homogeneous\\ KdV\\ equation\end{tabular} &
  \begin{tabular}[c]{@{}c@{}}Inhomogeneous\\ KdV\\ equation\end{tabular} \\ \hline
0\%   & 0  & 0  & 18 & -12 & 42 \\ \hline
25\%  & 0  & 6  & 22 & -4  & 32 \\ \hline
50\%  & 0  & -6 & 26 & 0   & 30 \\ \hline
75\%  & 0  & 14 & 2  & 6   & 20 \\ \hline
100\% & 10 & 10 & 0  & 6   & 10 \\ \hline
\end{tabular}
\end{table}

{
\section{MAE relative rates}
\label{sec:relative_maes}

For each type of equation a maximum MAE of all algorithms - PySINDy framework, modified and classical algorithms, is obtained. Then a mean value of each algorithm' MAE is normalized by the maximum value. The result can be considered as a relative mean absolute error, which allows us to compare the algorithms.

\begin{table}[]
\centering
\caption{Relative MAE for different noise levels (\%) - PySINDy framework. Green color represents the least relative error among all algorithms - PySINDy framework, modified algorithm with mf = 2.4, with tuned mixing factors and classical algorithm. Grey color denotes noise levels, where the algorithm could not obtain any equation.}
\label{tab:dfp_mae}
\begin{tabular}{|c|c|c|c|}
\hline
\begin{tabular}[c]{@{}c@{}}Noise \\ level\end{tabular} &
  \begin{tabular}[c]{@{}c@{}}Burgers'\\ equation\end{tabular} &
  \begin{tabular}[c]{@{}c@{}}Burgers'\\ equation\\ with\\ viscosity\end{tabular} &
  \begin{tabular}[c]{@{}c@{}}Homogeneous\\ KdV\\ equation\end{tabular} \\ \hline
0\%   & 99.95 & \cellcolor[HTML]{C5E0B3}0.01 & 0.44  \\ \hline
25\%  & 99.96 & 33.63                        & 1.15  \\ \hline
50\%  & 99.96 & \cellcolor[HTML]{C0C0C0}n/a  & 73.93 \\ \hline
75\%  & 99.96 & \cellcolor[HTML]{C0C0C0}n/a  & 91.96 \\ \hline
100\% & 99.96 & \cellcolor[HTML]{C0C0C0}n/a  & 99.39 \\ \hline
\end{tabular}
\end{table}

\begin{table}[]
\centering
\caption{Relative MAE for different noise levels (\%) - modified algorithm (mf = 2.4). Green color represents the least relative error among all algorithms - PySINDy framework, modified algorithm with mf = 2.4, with tuned mixing factors and classical algorithm. Grey color denotes noise levels, where the algorithm could not obtain any equation.}
\label{tab:dfs_mae}
\begin{tabular}{|c|c|c|c|}
\hline
\begin{tabular}[c]{@{}c@{}}Noise \\ level\end{tabular} &
  \begin{tabular}[c]{@{}c@{}}Burgers'\\ equation\end{tabular} &
  \begin{tabular}[c]{@{}c@{}}Burgers'\\ equation\\ with\\ viscosity\end{tabular} &
  \begin{tabular}[c]{@{}c@{}}Homogeneous\\ KdV\\ equation\end{tabular} \\ \hline
0\%   & \cellcolor[HTML]{C5E0B3}0.25 & 0.67                        & \cellcolor[HTML]{C5E0B3}0.11 \\ \hline
25\%  & \cellcolor[HTML]{C5E0B3}0.64 & 0.69                        & \cellcolor[HTML]{C5E0B3}0.11 \\ \hline
50\%  & 0.97                         & 1.95                        & \cellcolor[HTML]{C5E0B3}0.1  \\ \hline
75\%  & \cellcolor[HTML]{C5E0B3}0.89 & 1.8                         & \cellcolor[HTML]{C5E0B3}0.12 \\ \hline
100\% & \cellcolor[HTML]{C5E0B3}1.3  & \cellcolor[HTML]{C0C0C0}n/a & 0.43                         \\ \hline
\end{tabular}
\end{table}

\begin{table}[]
\centering
\caption{Relative MAE for different noise levels (\%) - modified algorithm with tuned mixing factors. Green color represents the least relative error among all algorithms - PySINDy framework, modified algorithm with mf = 2.4, with tuned mixing factors and classical algorithm.}
\label{tab:dfs_tuned_mae}
\begin{tabular}{|c|c|c|c|}
\hline
\begin{tabular}[c]{@{}c@{}}Noise \\ level\end{tabular} &
  \begin{tabular}[c]{@{}c@{}}Burgers'\\ equation\end{tabular} &
  \begin{tabular}[c]{@{}c@{}}Burgers'\\ equation\\ with\\ viscosity\end{tabular} &
  \begin{tabular}[c]{@{}c@{}}Homogeneous\\ KdV\\ equation\end{tabular} \\ \hline
0\%   & 0.32                         & 0.66                         & \cellcolor[HTML]{C5E0B3}0.10 \\ \hline
25\%  & 0.68                         & \cellcolor[HTML]{C5E0B3}0.68 & \cellcolor[HTML]{C5E0B3}0.10 \\ \hline
50\%  & 1.56                         & \cellcolor[HTML]{C5E0B3}1.32 & \cellcolor[HTML]{C5E0B3}0.10 \\ \hline
75\%  & 1.16                         & \cellcolor[HTML]{C5E0B3}1.7  & \cellcolor[HTML]{C5E0B3}0.10 \\ \hline
100\% & \cellcolor[HTML]{C5E0B3}1.32 & \cellcolor[HTML]{C5E0B3}2.92 & \cellcolor[HTML]{C5E0B3}0.34 \\ \hline
\end{tabular}
\end{table}

\begin{table}[]
\centering
\caption{Relative MAE for different noise levels (\%) - classical algorithm. Green color represents the least relative error among all algorithms - PySINDy framework, modified algorithm with mf = 2.4, with tuned mixing factors and classical algorithm. Grey color denotes noise levels, where the algorithm could not obtain any equation.}
\label{tab:dfo_mae}
\begin{tabular}{|c|c|c|c|}
\hline
\begin{tabular}[c]{@{}c@{}}Noise \\ level\end{tabular} &
  \begin{tabular}[c]{@{}c@{}}Burgers'\\ equation\end{tabular} &
  \begin{tabular}[c]{@{}c@{}}Burgers'\\ equation\\ with\\ viscosity\end{tabular} &
  \begin{tabular}[c]{@{}c@{}}Homogeneous\\ KdV\\ equation\end{tabular} \\ \hline
0\%   & 0.43                         & 0.67                         & \cellcolor[HTML]{C5E0B3}0.11 \\ \hline
25\%  & 1.16                         & \cellcolor[HTML]{C5E0B3}0.68 & \cellcolor[HTML]{C5E0B3}0.12 \\ \hline
50\%  & \cellcolor[HTML]{C5E0B3}0.93 & 2.4                          & \cellcolor[HTML]{C5E0B3}0.1  \\ \hline
75\%  & 1.15                         & \cellcolor[HTML]{C5E0B3}1.71 & \cellcolor[HTML]{C5E0B3}0.1  \\ \hline
100\% & \cellcolor[HTML]{C0C0C0}n/a  & \cellcolor[HTML]{C0C0C0}n/a  & \cellcolor[HTML]{C0C0C0}n/a  \\ \hline
\end{tabular}
\end{table}

}

{
\section{Algorithms hyper-parameters}
\label{sec:hyperparameters}

We note that both EPDE algorithms - classic and modified, share the same hyper-parameters in regards to the same type of input data and irrespective of the noise levels.
}

\begin{table}[h!]
\centering
\caption{Hyper-parameter values, EPDE algorithms}
\label{tab:hyperepde}
\begin{tabular}{|c|c|c|c|c|}
\hline
Hyper-parameter &
  \begin{tabular}[c]{@{}c@{}}Burgers'\\ equation\end{tabular} &
  \begin{tabular}[c]{@{}c@{}}Wave\\ equation\end{tabular} &
  \begin{tabular}[c]{@{}c@{}}Burgers' \\ equation\\ with viscosity\end{tabular} &
  \begin{tabular}[c]{@{}c@{}}Korteweg-de \\ Vries\\ equations\end{tabular} \\ \hline
Population size                                                                         & 5      & 5      & 8      & 8      \\ \hline
Training epochs                                                                         & 5      & 5      & 7      & 90     \\ \hline
\begin{tabular}[c]{@{}c@{}}Max number of \\ terms in equation\end{tabular}              & 3      & 3      & 3      & 4      \\ \hline
\begin{tabular}[c]{@{}c@{}}Max number of \\ factors in a term\end{tabular}              & 2      & 1      & 2      & 2      \\ \hline
\begin{tabular}[c]{@{}c@{}}Max derivative \\ order: \\ (by time, by space)\end{tabular} & (1, 1) & (2, 2) & (1, 2) & (1, 3) \\ \hline
\end{tabular}
\end{table}

\begin{table}[]
\centering
\caption{Hyper-parameter values, PySINDy package}
\label{tab:hyperpysindy}
\begin{tabular}{|c|c|c|c|}
\hline
Hyper-parameter &
  \begin{tabular}[c]{@{}c@{}}Burgers'\\ equation\\ with\\ viscosity\end{tabular} &
  \begin{tabular}[c]{@{}c@{}}Korteweg-de\\ Vries\\ equation\end{tabular} &
  \begin{tabular}[c]{@{}c@{}}Burgers'\\ equation\end{tabular} \\ \hline
Library functions                                                                     & $u, u^2$ & $u, u^2$ & $u, u^2$ \\ \hline
Derivative order                                                                      & 3        & 3        & 2        \\ \hline
Optimizer                                                                             & STLSQ    & SR3      & SSR      \\ \hline
\begin{tabular}[c]{@{}c@{}}Max number of \\ iterations\end{tabular}                   & 20       & 10000    & 20       \\ \hline
\begin{tabular}[c]{@{}c@{}}Threshold \\ (in coefficients\\ thresholding)\end{tabular} & 2        & 7        & -        \\ \hline
\end{tabular}
\end{table}

\newpage

\section{SymNet modifications for initial guess generation}
\label{app:modified_symnet}

A knowledge extraction approach is inspired by PDE-Net 2.0\footnote{https://github.com/ZichaoLong/PDE-Net/tree/PDE-Net-2.0} \cite{long2018pde,long2019pde}. In particular, it is also an equation discovery algorithm used to obtain the equations with a fixed structure, as shown in Eq.~\ref{eq:pdeproblem}. The authors proposed defining the structure $\Sigma'$ of the equation as a symbolic neural network, which in \cite{long2019pde} is called SymNet.

\begin{equation}
U_t = F(U, \nabla{U}, \nabla^2{U}, ...),  x \in \Omega \subset \mathbb{R}^2, t \in [0,T].
    \label{eq:pdeproblem}
\end{equation}

However, to retain possible structures $\Sigma'$ variety of the evolutionary algorithm, the initial SymNet architecture requires significant changes. After that, the initial guess of the equation is converted into the term importance, which is, in turn, used to guide structural optimization.

As was mentioned, Eq.~\ref{eq:pdeproblem} imposes certain restrictions onto the form of the equation. Therefore, the original algorithm was modified in such a way that the response function $F$ is dependent on temporal and spatial grids, as well as the time derivative, which does not act as a balance term on the left side of the equation (Eq.~\ref{eq:symnetproblem}).

\begin{equation}
    \left[\begin{matrix}
    U_t = F(t, x, U, U_x, U_{xx}, U_{tt}, U_{ttt}, ...),\\
    U_{tt} = F(t, x, U, U_x, U_{t}, U_{xx}, U_{ttt}, ...),\\
    U_{ttt} = F(t, x, U, U_x, U_{t}, U_{xx}, U_{tt}, ...),\\
    ...\\\end{matrix}\right.
    \label{eq:symnetproblem}
\end{equation}

\paragraph{SymNet architecture} 

\begin{figure}[ht!]
    \centering
    \includegraphics[width=0.9\linewidth]{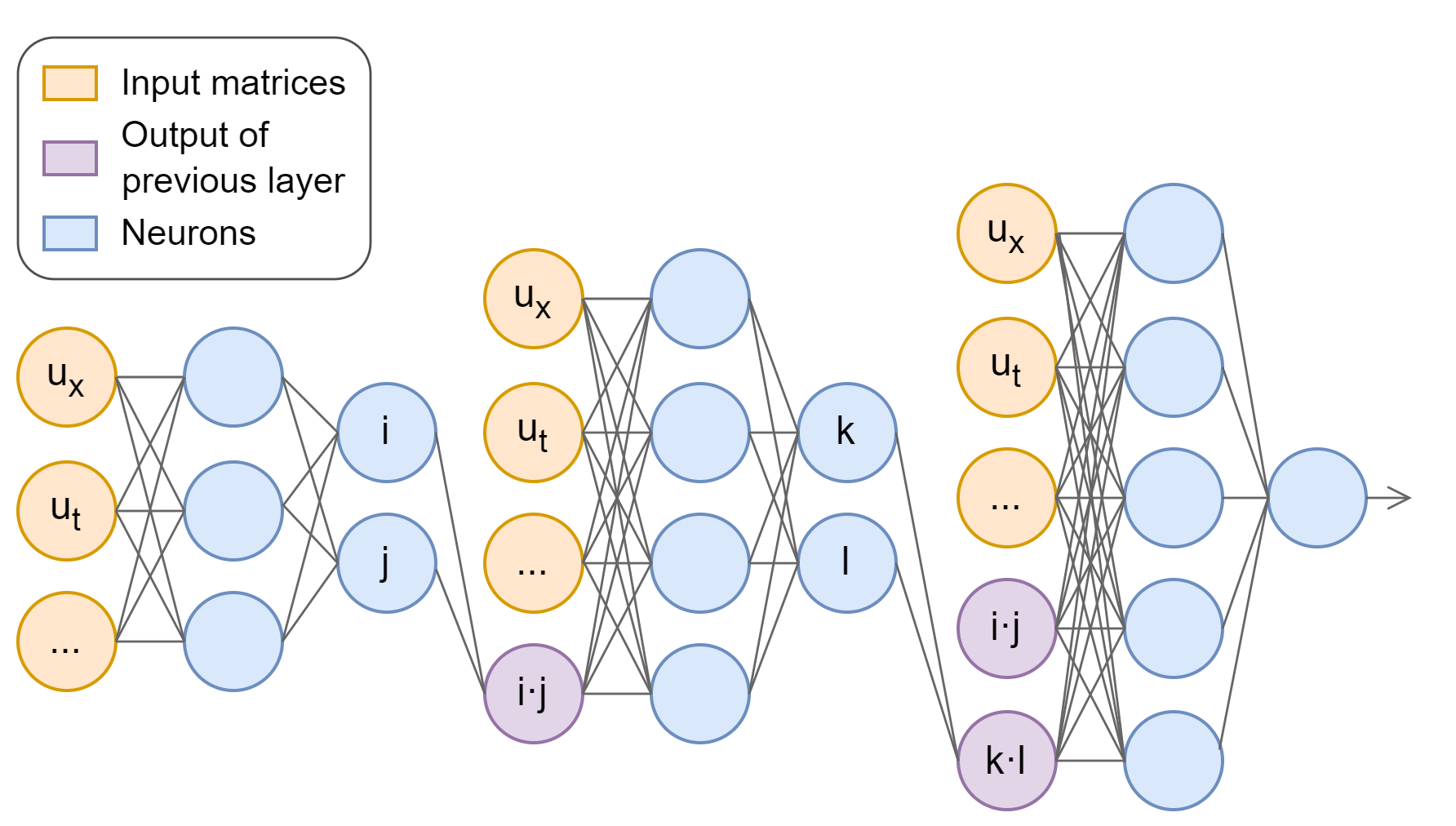}
    \caption{The schematic diagram of SymNet - example with one hidden layer.}
    \label{fig:symnet}
\end{figure}

The schematic modified SymNet architecture is shown in Fig.~\ref{fig:symnet}. For simplicity of understanding, only one hidden layer is illustrated. However, we note that there could be a multi-layer SymNet.

The input of SymNet is a set of tokens (Eq.~\ref{eq:token}) pre-computed on an observation grid. The output is a response function $F$ value computed on an observation grid. The main modification is the possibility of working in a multi-layer mode with arbitrary tokens (apart from spatial derivatives as in classical SymNet). The maximum order of the resulting polynomials is equal to the number of hidden layers incremented by one. For example, the maximum order of polynomials computed with the model on Fig.~\ref{fig:symnet} is 2. 

When the model is trained, the layers' parameters can be used to obtain the symbolic representation of the equation. For this goal, symbolic token labels are inputted into the SymNet structure with weights identified within the training procedure.

\paragraph{Loss function and regularization}

Afterward, every generated architecture from Eq.~\ref{eq:symnetproblem} is trained using gradient methods with the loss function based on the regularized loss function proposed in \cite{long2019pde} in the form shown in Eq.~\ref{eq:symnetloss}.

\begin{equation}
    L = L^{data} + \lambda L^{SymNet},
    \label{eq:symnetloss}
\end{equation}

{In Eq.~\ref{eq:symnetloss} hyper-parameter $\lambda$ is chosen to be either 0.001 or $10^{-7}$}, $L^{data}$ is a loss part that depends on the initial observation data: it measures the discrepancy of the SymNet right-hand side output $\widetilde{F}$ from the left-hand side temporal derivative $\widetilde{U}$ computed at the entire observational data grid $X$. The exact formulation of $L^{data}$ is given in Eq.~\ref{eq:symnetloss1}.

\begin{equation}
\begin{array}{cc}
    L^{data} = \Big[||\widetilde{U}-\widetilde{F}||_2^2  \Big]_X
    %=\frac{1}{N\cdot M}\sum\limits_{i=1}^{N}{\sum\limits_{j=1}^{M}{||U_{Tij}-\widetilde{U}_{ij}||_2^2}}, \\ x = x_1, ..., x_i, ..., x_N, t = t_1, ..., t_j, ..., t_M.
\end{array}
    \label{eq:symnetloss1}
\end{equation}

The term $L^{SymNet}$ is a regularization loss term for SymNet. It is based on Huber's loss function $l_1^s$, where the threshold is equal to $0.001$. The term $L^{SymNet}$ has the form Eq.~\ref{eq:symnetloss2}.

\begin{equation}
\begin{array}{cc}
    L^{SymNet} = \sum\limits_{w \in parameters \; of \; SymNet}{l_1^s(w)}; \\[15pt]
    l_1^s(w) = 
    \begin{cases}
        |w| - \frac{s}{2} \quad if \; |w| > s, \\
        \frac{1}{2s}w^2 \quad \text{else}.
    \end{cases} 
\end{array}
    \label{eq:symnetloss2}
\end{equation}

{SymNet training procedure is done for every equation of the system~\ref{eq:symnetproblem} for two possible values of $\lambda$.} The equation with the minimal loss function Eq.~\ref{eq:symnetloss} is chosen to construct the term importance distribution.

\newpage

\section{Initial-boundary value problem statements}
\label{app:IVP_statements}

\subsection{Wave equation}
\label{app:wave_eq}

The initial-boundary value problem for the wave equation is given in Eq.~\ref{eq:wave_eq}.

\begin{equation}
\begin{array}{cc}
\frac{\partial^2 u}{\partial t^2} - \frac{1}{25} \frac{\partial^2 u}{\partial x^2} = 0 \\
u(0,t)=u(1,t)=0\\
u(x,0)=10^4 \sin^2{\frac{1}{10}x(x-1)}\\
u'(x,0)=10^3 \sin^2{\frac{1}{10}x(x-1)}\\
(x,t) \in [0,1] \times [0,1]
\end{array}
    \label{eq:wave_eq}
\end{equation}

The solution of Eq.~\ref{eq:wave_eq} was obtained with the Wolfram Mathematica software interpolation method on the grid of $101 \times 101$ discretization points in the domain $(x,t) \in [0,1] \times [0,1]$.

\subsection{Korteweg-de Vries equation}
\label{app:kdv_eq}

The initial-boundary value problem is given in Eq.~\ref{eq:KdV}. The solution of the equation was obtained in the same manner as in the case of a wave equation - with the aid of Wolfram Mathematica software. The grid consists of $101 \times 101$ discretization points in the domain $(x,t) \in [0,1] \times [0,1]$. The derivatives were obtained by differentiating the interpolated solution within Wolfram Mathematica software.

\begin{equation}
\begin{array}{cc}
u_t + 6 u u_x + u_{xxx} = \cos t \sin x \\
u(x,0)=0\\
\left[u_{xx}+2 u_x+u \right] \Big|_{x=0}=0\\
\left[2 u_{xx}+u_x+3 u \right] \Big|_{x=1}=0\\
\left[5 u_x+5 u \right] \Big|_{x=1}=0\\
(x,t) \in [0,1] \times [0,1]
\end{array}
    \label{eq:KdV}
\end{equation}

\subsection{Inviscid Burger's equation}
\label{app:burg_eq}

The initial-boundary value problem for Burger's equation is represented with Eq.~\ref{eq:burgers_eq}.

\begin{equation}
\begin{array}{cc}
     \frac{\partial u}{\partial t} + u \frac{\partial u}{\partial x} = 0  \\
     u(0,t)=\begin{cases}
    1000, t\geq 2 \\
    0, t < 2
\end{cases}\\
u(x,0)=\begin{cases}
    1000, x \leq - 2000 \\
    -x/2, -2000<x<0 \\
    0, \text{otherwise}
\end{cases}\\
(x,t) \in [-4000,4000] \times [0,4]
\end{array}
    \label{eq:burgers_eq}
\end{equation}

The analytical solution to the problem presented in Eq.~\ref{eq:burgers_eq} is given in \cite{rudy2017data}. Data for the experiment were obtained with the discretization of the solution in the domain $(x,t) \in [-4000,4000] \times [0,4]$ using $101 \times 101$ points.

\subsection{Burger's equation with viscosity}
\label{app:burgvisc_eq}

The problem and data were provided by the authors of PySINDY\footnote{https://github.com/dynamicslab/pysindy} \cite{desilva2020,Kaptanoglu2022}. The problem can be formulated in Eq.~\ref{eq:Burgers2}, where the boundary conditions were not reported. The solution was provided for the domain $(x,t) \in [-8,8] \times [0,10]$ using $ 256 \times 101$ discretization points.

\begin{equation}
\begin{array}{cc}
\frac{\partial u}{\partial t} + u \frac{\partial u}{\partial x} - 0.1 \frac{\partial^2 u}{\partial x^2} = 0 \\
(x,t) \in [-8,8] \times [0,10]
\end {array}
    \label{eq:Burgers2}
\end{equation}

\subsection{PYSINDy Korteweg -- de Vries equation}
\label{app:homokdv_eq}

As in the case of Burgers' equation, the data and the problem (Eq.~\ref{eq:KdV_homo}) were provided by the authors of PySINDY \footnote{https://github.com/dynamicslab/pysindy} \cite{desilva2020,Kaptanoglu2022} for the domain $(x,t) \in [-30,30] \times [0,20]$ using $ 512 \times 201$ discretization points.

\begin{equation}
\begin{array}{cc}
\frac{\partial u}{\partial t} + 6 u \frac{\partial u}{\partial x} + \frac{\partial^3 u}{\partial x^3} = 0 \\
(x,t) \in [-30,30] \times [0,20]
\end {array}
    \label{eq:KdV_homo}
\end{equation}

\section{Resulting Symnet equations}
\label{app:symnet_results}

\paragraph{Burgers' equation}

The setup of the experiment follows the one from ~\ref{app:burg_eq}. The performance metrics are presented in Tab. ~\ref{tab:symnet_mae} and Tab. ~\ref{tab:symnet_shd}. Although the MAE value is not high, the SHD value is fairly large. An example of discovered equation is given in Eq. ~\ref{eq:burg1_symalone}. The total number of terms in this equation is equal to 34 with the precision of 16 digits.

\begin{equation}
\begin{aligned}
\frac{\partial u}{\partial t} = -1.025 u \cdot \frac{\partial u}{\partial x} + 0.188 \frac{\partial u}{\partial x} -0.111 u \cdot \frac{\partial^2 u}{\partial x^2} + 0.089 \frac{\partial^2 u}{\partial x^2} \:+\\+\: 0.033 -0.022 (\frac{\partial u}{\partial x})^2 -0.015 (\frac{\partial u}{\partial x})^2 \cdot u + ... \: (total \: 34 \: terms)
\end{aligned}
\label{eq:burg1_symalone}
\end{equation}

\paragraph{Burgers' equation with viscosity term}

The problem for this type of experiment is formulated in ~\ref{app:burgvisc_eq}. Similarly to the previous experiment, the SHD and  MAE metrics are given in Tab. ~\ref{tab:symnet_shd}, ~\ref{tab:symnet_mae}; and the example of discovered equation is presented below (Eq. ~\ref{eq:burg2_symalone}). The algorithm was able to find both terms that compose the Burgers' equation with viscosity term, however, there are several extra terms, the value of which is comparable to that of the viscosity term, although, overall MAE metric is relatively small.

\begin{equation}
\begin{aligned}
\frac{\partial u}{\partial t} = -0.971 u \frac{\partial u}{\partial x} + 0.096 \frac{\partial^2 u}{\partial x^2} - 0.035 u \frac{\partial^2 u}{\partial x^2} + \\
+ 0.034 u^2 \frac{\partial u}{\partial x} - 0.031 \frac{\partial u}{\partial x} \cdot \frac{\partial^2 u}{\partial x^2} - 0.028 u^2 - 0.019 u^3 + \\
+ 0.016 u^2 \frac{\partial^2 u}{\partial x^2} + 0.009 u - 0.005 \frac{\partial u}{\partial x} + 0.003 u \frac{\partial u}{\partial x} \cdot \frac{\partial^2 u}{\partial x^2} + \\ + ... \: (total \: 35 \: terms)
\end{aligned}
    \label{eq:burg2_symalone}
\end{equation}

\paragraph{Wave equation}

The problem of wave equation is defined in ~\ref{app:wave_eq}. The metrics in tables ~\ref{tab:symnet_mae} and  ~\ref{tab:symnet_shd} does not differ much from the previous experiments. On the other hand, the form of the discovered equations (example in Eq.\ref{eq:wave_symalone}) does not correspond to the desired structure. We note that with the noise increase in the data the form of found equation changes drastically, due to the fact that the algorithm is attempting to balance the first time derivative and not the second one.

\begin{equation}
\begin{aligned}
\frac{\partial^2 u}{\partial t^2} = 1.351 \frac{\partial u}{\partial t} -1.037 + 0.516 (u)^2 -0.407 (\frac{\partial u}{\partial t})^2 -0.331 u \cdot \frac{\partial u}{\partial t} - 0.324 u \\ -0.104 (u)^3 + 0.076 (\frac{\partial u}{\partial t})^2 \cdot u + 0.029 (\frac{\partial u}{\partial t})^3 + 0.024 u \cdot \frac{\partial^2 u}{\partial x^2} -0.021 (u)^2 \cdot \frac{\partial u}{\partial t} \\ + 0.017 \frac{\partial^2 u}{\partial x^2}  + ...  \: (total \: 70 \: terms)
\end{aligned}
    \label{eq:wave_symalone}
\end{equation}

\paragraph{Inhomogeneous Korteweg -- de Vries equation}

Experimental setup is given in ~\ref{app:kdv_eq}. With the results presented in Tab.~\ref{tab:symnet_mae}, ~\ref{tab:symnet_shd} and Eq.~\ref{eq:kdvin_symalone} we can conclude that more complex cases of partial differential equations are fairly challenging for the SymNet algorithm alone. Apparently, the difference in coefficients with the ground truth equation is reasonably high; the number of terms, where the coefficient values are larger than $10^{-6}$ is also often over 100.

\begin{equation}
\begin{aligned}
\frac{\partial u}{\partial t} = -2.115 u \cdot \frac{\partial u}{\partial x} + 1.045 (u)^2 + 1.03 cos(t)sin(x) -0.988 \frac{\partial^3 u}{\partial x^3} + \\ + 0.905 (\frac{\partial u}{\partial x})^2 + 0.867 u \cdot \frac{\partial u}{\partial x} \cdot \frac{\partial^2 u}{\partial x^2} -0.834 (\frac{\partial u}{\partial x})^2 \cdot u + ... \: (total \: 126 \: terms)
\end{aligned}
    \label{eq:kdvin_symalone}
\end{equation}

\paragraph{Korteweg -- de Vries equation}
The problem was formulated in ~\ref{app:homokdv_eq}. The results in tables ~\ref{tab:symnet_mae}, ~\ref{tab:symnet_shd} and Eq.~\ref{eq:kdv_symalone} indicate that the algorithm can not obtain the desired equation and the coefficients error of the discovered equation is significantly high.

\begin{equation}
\label{eq:kdv_symalone}
\begin{aligned}
\frac{\partial u}{\partial t} = -0.891 \frac{\partial u}{\partial x} + 0.035 \frac{\partial^3 u}{\partial x^3} -0.004 \frac{\partial u}{\partial x} \cdot \frac{\partial^2 u}{\partial x^2} + 0.004 u \cdot \frac{\partial^2 u}{\partial x^2}- \\ -0.002 (\frac{\partial^2 u}{\partial x^2})^2 -0.002 \frac{\partial^3 u}{\partial x^3} \cdot \frac{\partial^2 u}{\partial x^2} -0.001 \frac{\partial^2 u}{\partial x^2} + ... \: (total \: 70 \: terms)
\end{aligned}
\end{equation}

\newpage

\section{Mixing factor sensitivity tables}
\label{app:mixing_factor_tables}

\begin{table}[h!]
\centering
\caption{Number of successful runs of modified algorithm (with classical as a benchmark) for different noise levels and mixing factors. Mixing factors by KL are 1.4 for all noise levels.}
\label{tab:mix_kdv_s}
\begin{tabular}{|cccccc|}
\hline
\multicolumn{6}{|c|}{Inhomogeneous KdV equation} \\ \hline
\multicolumn{1}{|c|}{\begin{tabular}[c]{@{}c@{}}mixing\\ factor\end{tabular}} &
  \multicolumn{1}{c|}{\begin{tabular}[c]{@{}c@{}}0\%\\ noise\end{tabular}} &
  \multicolumn{1}{c|}{\begin{tabular}[c]{@{}c@{}}25\%\\ noise\end{tabular}} &
  \multicolumn{1}{c|}{\begin{tabular}[c]{@{}c@{}}50\%\\ noise\end{tabular}} &
  \multicolumn{1}{c|}{\begin{tabular}[c]{@{}c@{}}75\%\\ noise\end{tabular}} &
  \begin{tabular}[c]{@{}c@{}}100\%\\ noise\end{tabular} \\ \hline
\multicolumn{1}{|c|}{2.4} &
  \multicolumn{1}{c|}{26} &
  \multicolumn{1}{c|}{27} &
  \multicolumn{1}{c|}{26} &
  \multicolumn{1}{c|}{28} &
  3 \\ \hline
\multicolumn{1}{|c|}{3.0} &
  \multicolumn{1}{c|}{24} &
  \multicolumn{1}{c|}{\cellcolor[HTML]{9ECA80}34} &
  \multicolumn{1}{c|}{25} &
  \multicolumn{1}{c|}{24} &
  3 \\ \hline
\multicolumn{1}{|c|}{3.6} &
  \multicolumn{1}{c|}{30} &
  \multicolumn{1}{c|}{30} &
  \multicolumn{1}{c|}{26} &
  \multicolumn{1}{c|}{28} &
  4 \\ \hline
\multicolumn{1}{|c|}{4.5} &
  \multicolumn{1}{c|}{25} &
  \multicolumn{1}{c|}{27} &
  \multicolumn{1}{c|}{27} &
  \multicolumn{1}{c|}{22} &
  \cellcolor[HTML]{9ECA80}9 \\ \hline
\multicolumn{1}{|c|}{by KL} &
  \multicolumn{1}{c|}{\cellcolor[HTML]{9ECA80}35} &
  \multicolumn{1}{c|}{25} &
  \multicolumn{1}{c|}{\cellcolor[HTML]{9ECA80}36} &
  \multicolumn{1}{c|}{\cellcolor[HTML]{9ECA80}30} &
  3 \\ \hline
\multicolumn{1}{|c|}{\begin{tabular}[c]{@{}c@{}}classical\\ alg.\end{tabular}} &
  \multicolumn{1}{c|}{30} &
  \multicolumn{1}{c|}{29} &
  \multicolumn{1}{c|}{26} &
  \multicolumn{1}{c|}{25} &
  0 \\ \hline
\end{tabular}
\end{table}

\begin{table}[h!]
\centering
\caption{Number of successful runs of modified algorithm (with classical as a benchmark) for different noise levels and mixing factors. Mixing factors by KL: 3.5, 3.4, 3.4, 3.5, 3.5.}
\label{tab:mix_burg}
\begin{tabular}{|cccccc|}
\hline
\multicolumn{6}{|c|}{Burgers' equation} \\ \hline
\multicolumn{1}{|c|}{\begin{tabular}[c]{@{}c@{}}mixing\\ factor\end{tabular}} &
  \multicolumn{1}{c|}{\begin{tabular}[c]{@{}c@{}}0\%\\ noise\end{tabular}} &
  \multicolumn{1}{c|}{\begin{tabular}[c]{@{}c@{}}25\%\\ noise\end{tabular}} &
  \multicolumn{1}{c|}{\begin{tabular}[c]{@{}c@{}}50\%\\ noise\end{tabular}} &
  \multicolumn{1}{c|}{\begin{tabular}[c]{@{}c@{}}75\%\\ noise\end{tabular}} &
  \begin{tabular}[c]{@{}c@{}}100\%\\ noise\end{tabular} \\ \hline
\multicolumn{1}{|c|}{2.4} &
  \multicolumn{1}{c|}{50} &
  \multicolumn{1}{c|}{\cellcolor[HTML]{9ECA80}49} &
  \multicolumn{1}{c|}{9} &
  \multicolumn{1}{c|}{\cellcolor[HTML]{9ECA80}18} &
  5 \\ \hline
\multicolumn{1}{|c|}{3.0} &
  \multicolumn{1}{c|}{50} &
  \multicolumn{1}{c|}{\cellcolor[HTML]{9ECA80}48} &
  \multicolumn{1}{c|}{14} &
  \multicolumn{1}{c|}{13} &
  7 \\ \hline
\multicolumn{1}{|c|}{3.6} &
  \multicolumn{1}{c|}{50} &
  \multicolumn{1}{c|}{46} &
  \multicolumn{1}{c|}{\cellcolor[HTML]{9ECA80}17} &
  \multicolumn{1}{c|}{16} &
  \cellcolor[HTML]{9ECA80}9 \\ \hline
\multicolumn{1}{|c|}{by KL} &
  \multicolumn{1}{c|}{50} &
  \multicolumn{1}{c|}{\cellcolor[HTML]{9ECA80}48} &
  \multicolumn{1}{c|}{\cellcolor[HTML]{9ECA80}17} &
  \multicolumn{1}{c|}{11} &
  \cellcolor[HTML]{9ECA80}8 \\ \hline
\multicolumn{1}{|c|}{\begin{tabular}[c]{@{}c@{}}classical\\ alg.\end{tabular}} &
  \multicolumn{1}{c|}{50} &
  \multicolumn{1}{c|}{46} &
  \multicolumn{1}{c|}{12} &
  \multicolumn{1}{c|}{11} &
  0 \\ \hline
\end{tabular}
\end{table}

\begin{table}[h!]
\centering
\caption{Number of successful runs of modified algorithm (with classical as a benchmark) for different noise levels and mixing factors. Mixing factors by KL are 3.2, 3.2, 3.1, 2.8, 2.8.}
\label{tab:mix_burg_s}
\begin{tabular}{|cccccc|}
\hline
\multicolumn{6}{|c|}{Burgers' equation with viscosity} \\ \hline
\multicolumn{1}{|c|}{\begin{tabular}[c]{@{}c@{}}mixing\\ factor\end{tabular}} &
  \multicolumn{1}{c|}{\begin{tabular}[c]{@{}c@{}}0\%\\ noise\end{tabular}} &
  \multicolumn{1}{c|}{\begin{tabular}[c]{@{}c@{}}25\%\\ noise\end{tabular}} &
  \multicolumn{1}{c|}{\begin{tabular}[c]{@{}c@{}}50\%\\ noise\end{tabular}} &
  \multicolumn{1}{c|}{\begin{tabular}[c]{@{}c@{}}75\%\\ noise\end{tabular}} &
  \begin{tabular}[c]{@{}c@{}}100\%\\ noise\end{tabular} \\ \hline
\multicolumn{1}{|c|}{2.4} &
  \multicolumn{1}{c|}{\cellcolor[HTML]{9ECA80}49} &
  \multicolumn{1}{c|}{48} &
  \multicolumn{1}{c|}{19} &
  \multicolumn{1}{c|}{3} &
  0 \\ \hline
\multicolumn{1}{|c|}{3.0} &
  \multicolumn{1}{c|}{\cellcolor[HTML]{9ECA80}48} &
  \multicolumn{1}{c|}{48} &
  \multicolumn{1}{c|}{\cellcolor[HTML]{9ECA80}{\color[HTML]{000000} 27}} &
  \multicolumn{1}{c|}{4} &
  0 \\ \hline
\multicolumn{1}{|c|}{3.6} &
  \multicolumn{1}{c|}{\cellcolor[HTML]{9ECA80}49} &
  \multicolumn{1}{c|}{\cellcolor[HTML]{9ECA80}50} &
  \multicolumn{1}{c|}{20} &
  \multicolumn{1}{c|}{4} &
  \cellcolor[HTML]{9ECA80}2 \\ \hline
\multicolumn{1}{|c|}{by KL} &
  \multicolumn{1}{c|}{46} &
  \multicolumn{1}{c|}{\cellcolor[HTML]{9ECA80}49} &
  \multicolumn{1}{c|}{20} &
  \multicolumn{1}{c|}{\cellcolor[HTML]{9ECA80}7} &
  \cellcolor[HTML]{9ECA80}1 \\ \hline
\multicolumn{1}{|c|}{\begin{tabular}[c]{@{}c@{}}classical\\ alg.\end{tabular}} &
  \multicolumn{1}{c|}{40} &
  \multicolumn{1}{c|}{37} &
  \multicolumn{1}{c|}{6} &
  \multicolumn{1}{c|}{2} &
  0 \\ \hline
\end{tabular}
\end{table}

\begin{table}[h!]
\centering
\caption{Number of successful runs of modified algorithm (with classical as a benchmark) for different noise levels and mixing factors. Mixing factors by KL are 4.7, 4.6, 4.5, 4.5, 4.1.}
\label{tab:mix_kdv}
\begin{tabular}{|cccccc|}
\hline
\multicolumn{6}{|c|}{Homogeneous KdV equation} \\ \hline
\multicolumn{1}{|c|}{\begin{tabular}[c]{@{}c@{}}mixing\\ factor\end{tabular}} &
  \multicolumn{1}{c|}{\begin{tabular}[c]{@{}c@{}}0\%\\ noise\end{tabular}} &
  \multicolumn{1}{c|}{\begin{tabular}[c]{@{}c@{}}25\%\\ noise\end{tabular}} &
  \multicolumn{1}{c|}{\begin{tabular}[c]{@{}c@{}}50\%\\ noise\end{tabular}} &
  \multicolumn{1}{c|}{\begin{tabular}[c]{@{}c@{}}75\%\\ noise\end{tabular}} &
  \begin{tabular}[c]{@{}c@{}}100\%\\ noise\end{tabular} \\ \hline
\multicolumn{1}{|c|}{2.4} &
  \multicolumn{1}{c|}{34} &
  \multicolumn{1}{c|}{35} &
  \multicolumn{1}{c|}{28} &
  \multicolumn{1}{c|}{20} &
  5 \\ \hline
\multicolumn{1}{|c|}{3.0} &
  \multicolumn{1}{c|}{\cellcolor[HTML]{9ECA80}48} &
  \multicolumn{1}{c|}{34} &
  \multicolumn{1}{c|}{26} &
  \multicolumn{1}{c|}{29} &
  5 \\ \hline
\multicolumn{1}{|c|}{by KL} &
  \multicolumn{1}{c|}{\cellcolor[HTML]{9ECA80}49} &
  \multicolumn{1}{c|}{\cellcolor[HTML]{9ECA80}42} &
  \multicolumn{1}{c|}{\cellcolor[HTML]{9ECA80}42} &
  \multicolumn{1}{c|}{\cellcolor[HTML]{9ECA80}34} &
  \cellcolor[HTML]{9ECA80}6 \\ \hline
\multicolumn{1}{|c|}{\begin{tabular}[c]{@{}c@{}}classical\\ alg.\end{tabular}} &
  \multicolumn{1}{c|}{13} &
  \multicolumn{1}{c|}{19} &
  \multicolumn{1}{c|}{13} &
  \multicolumn{1}{c|}{10} &
  0 \\ \hline
\end{tabular}
\end{table}

\begin{table}[h!]
\centering
\caption{Number of successful runs of modified algorithm (with classical as a benchmark) for different noise levels and mixing factors. Mixing factors by KL:}
\label{tab:mix_wave}
\begin{tabular}{|cccccc|}
\hline
\multicolumn{6}{|c|}{Wave equation} \\ \hline
\multicolumn{1}{|c|}{\begin{tabular}[c]{@{}c@{}}mixing\\ factor\end{tabular}} &
  \multicolumn{1}{c|}{\begin{tabular}[c]{@{}c@{}}0\%\\ noise\end{tabular}} &
  \multicolumn{1}{c|}{\begin{tabular}[c]{@{}c@{}}25\%\\ noise\end{tabular}} &
  \multicolumn{1}{c|}{\begin{tabular}[c]{@{}c@{}}50\%\\ noise\end{tabular}} &
  \multicolumn{1}{c|}{\begin{tabular}[c]{@{}c@{}}75\%\\ noise\end{tabular}} &
  \begin{tabular}[c]{@{}c@{}}100\%\\ noise\end{tabular} \\ \hline
\multicolumn{1}{|c|}{2.4} &
  \multicolumn{1}{c|}{50} &
  \multicolumn{1}{c|}{50} &
  \multicolumn{1}{c|}{50} &
  \multicolumn{1}{c|}{50} &
  \cellcolor[HTML]{9ECA80}5 \\ \hline
\multicolumn{1}{|c|}{by KL} &
  \multicolumn{1}{c|}{50} &
  \multicolumn{1}{c|}{50} &
  \multicolumn{1}{c|}{50} &
  \multicolumn{1}{c|}{50} &
  3 \\ \hline
\multicolumn{1}{|c|}{\begin{tabular}[c]{@{}c@{}}classical\\ alg.\end{tabular}} &
  \multicolumn{1}{c|}{50} &
  \multicolumn{1}{c|}{50} &
  \multicolumn{1}{c|}{50} &
  \multicolumn{1}{c|}{50} &
  0 \\ \hline
\end{tabular}
\end{table}

\newpage

\section{Time consumption experiments}
\label{app:time_consumption}

\begin{figure}[ht!]
    \centering
    \includegraphics[width=0.9\linewidth]{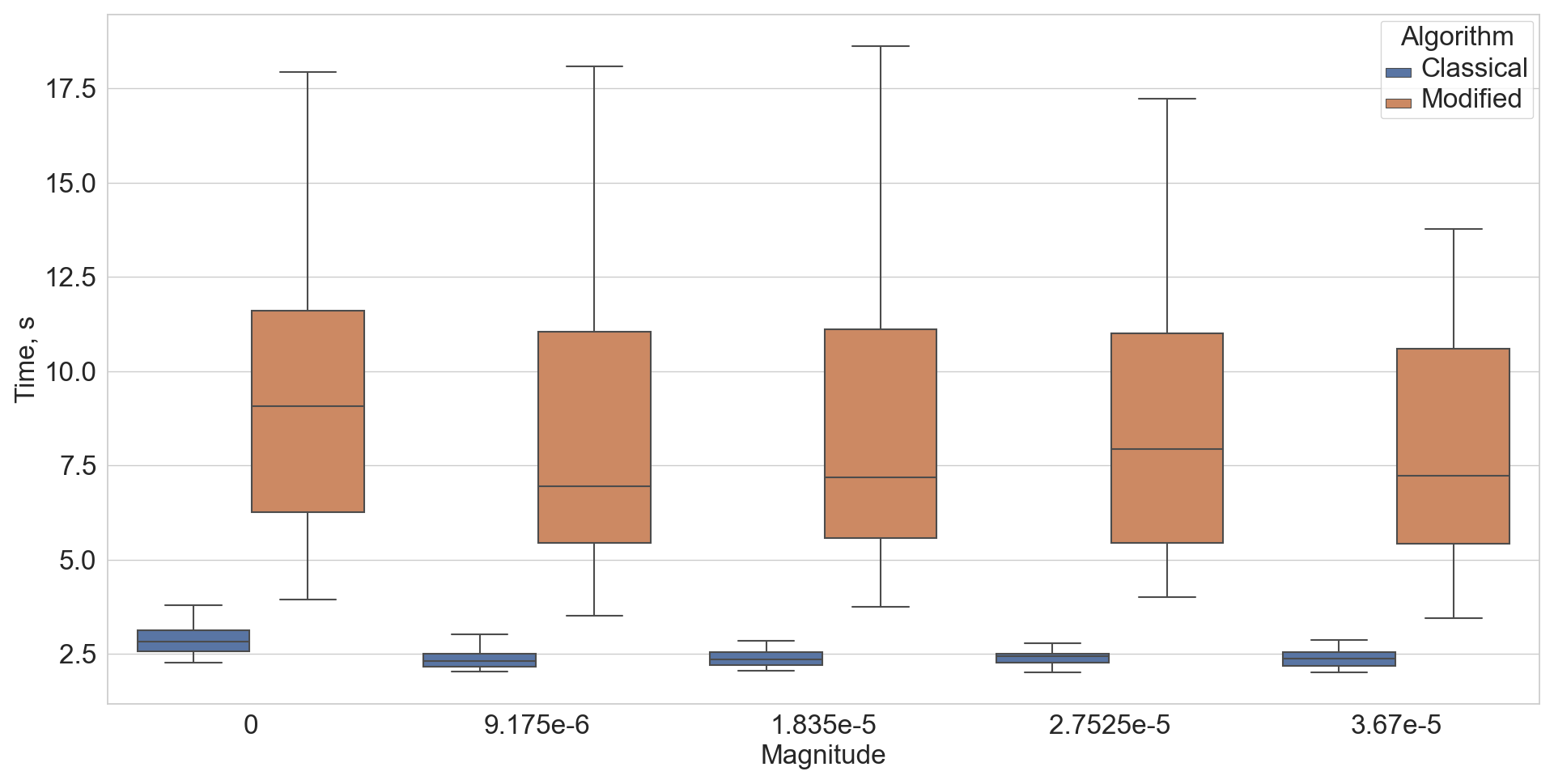}
    \caption{Algorithms running time for different noise magnitude values, Burgers' equation. The average running time of the PySINDY algorithm through all magnitudes is around 0.021 s.}
    \label{fig:burgers_time}
\end{figure}

\begin{figure}[ht!]
    \centering
    \includegraphics[width=0.9\linewidth]{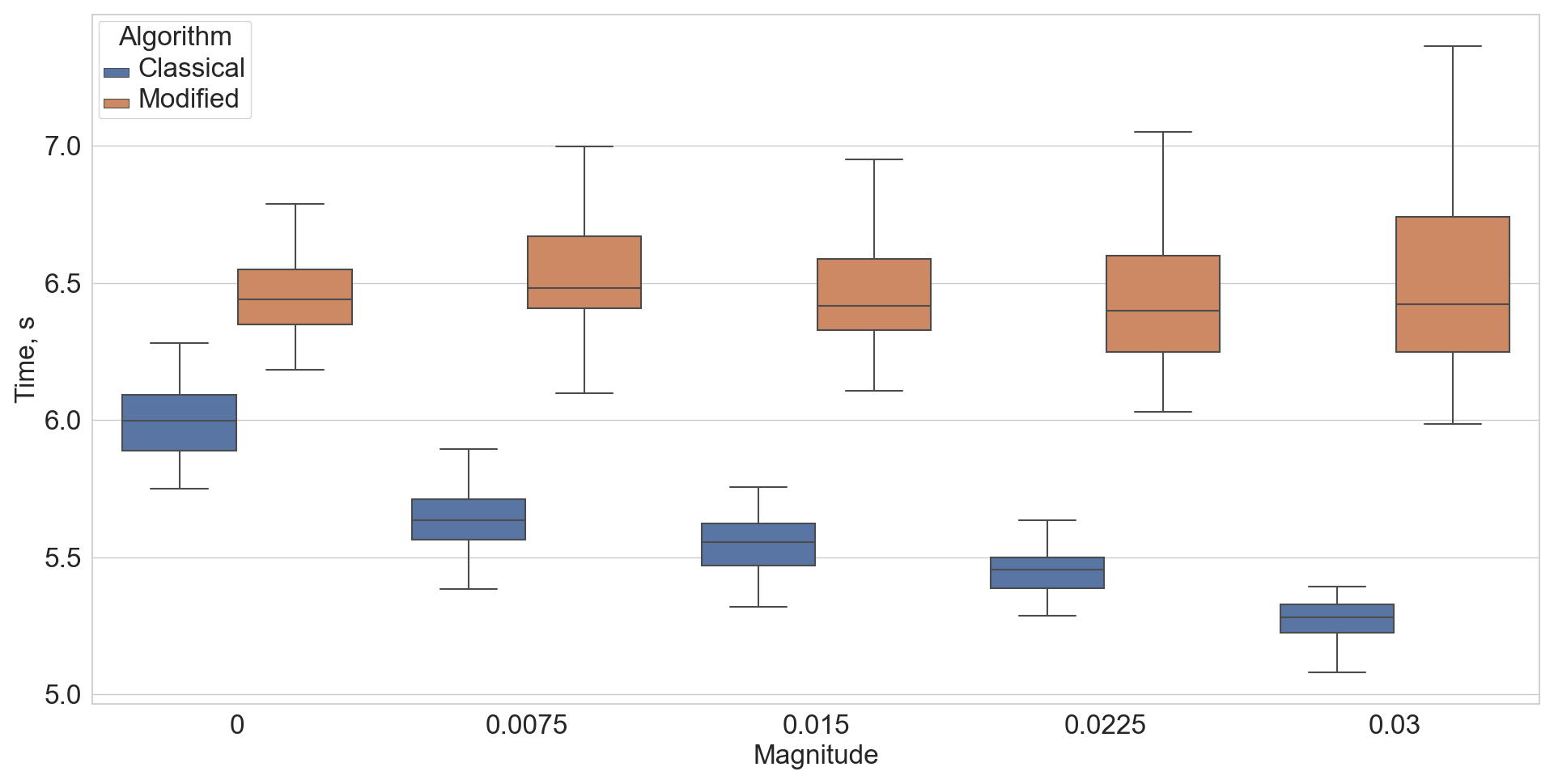}
    \caption{Algorithms running time for different noise magnitude values, Burgers' equation. The average running time of the PySINDY algorithm through all magnitudes is around 0.01 s. }
    \label{fig:burgers_s_time}
\end{figure}

\begin{figure}[ht!]
    \centering
    \includegraphics[width=0.9\linewidth]{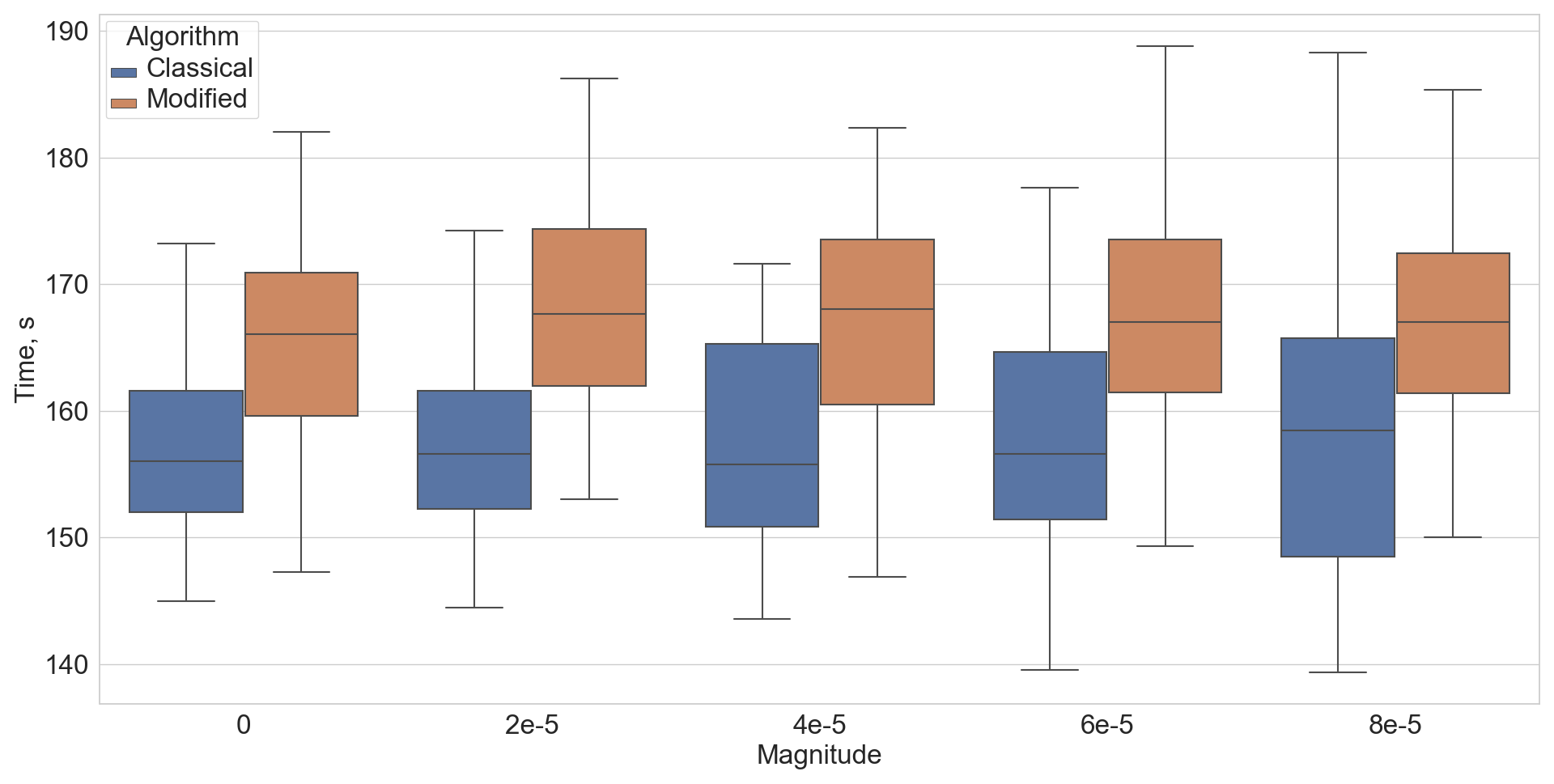}
    \caption{Algorithms running time for different noise magnitude values, Korteweg -- de Vries equation. The average running time of the PySINDY algorithm through all magnitudes is around 0.08 s.}
    \label{fig:kdv_s_time}
\end{figure}

\begin{figure}[ht!]
    \centering
    \includegraphics[width=0.9\linewidth]{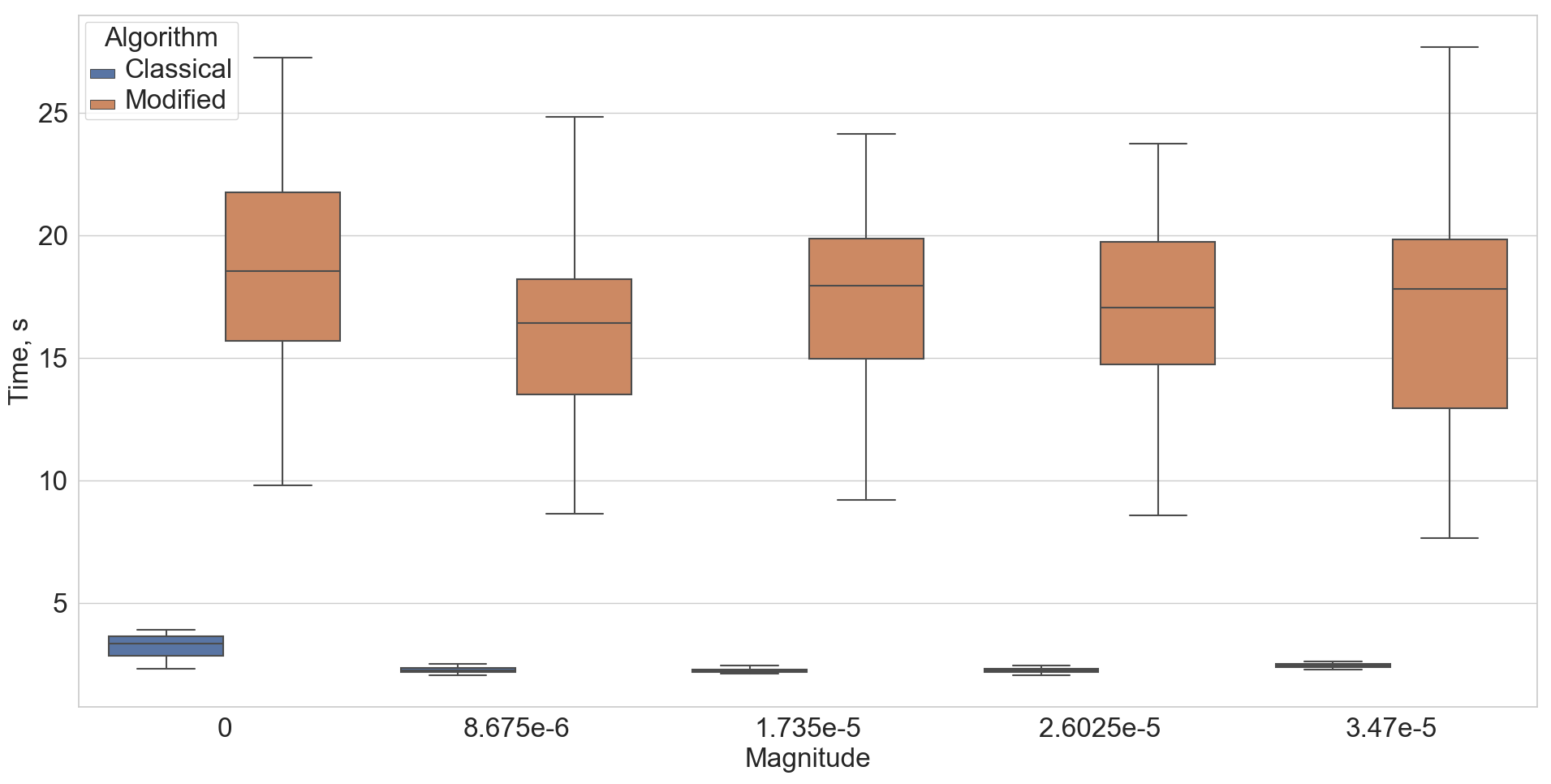}
    \caption{Algorithms running time for different noise magnitude values, Wave equation.}
    \label{fig:wave_time}
\end{figure}

\begin{figure}[ht!]
    \centering
    \includegraphics[width=0.9\linewidth]{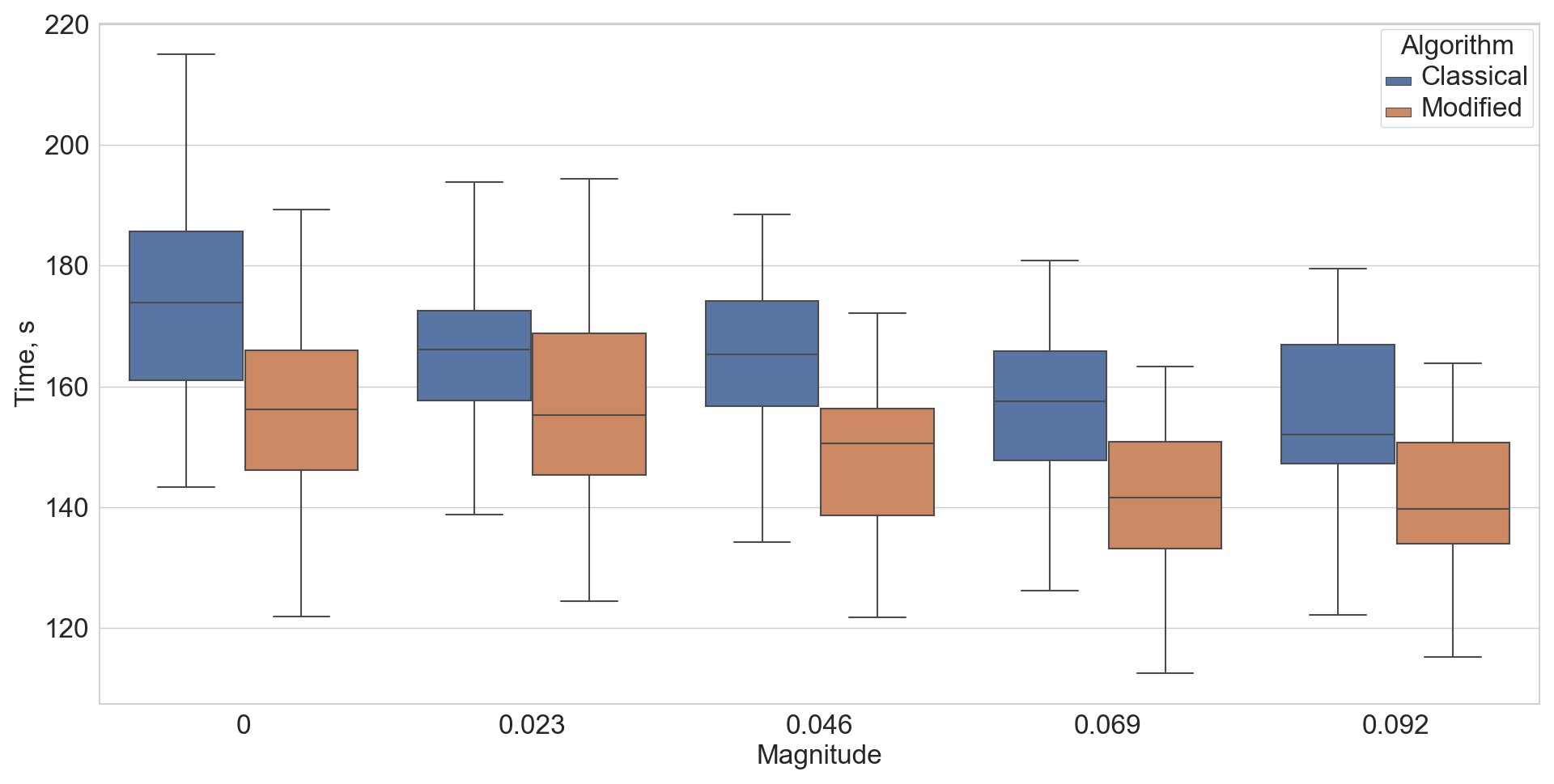}
    \caption{Algorithms running time for different noise magnitude values, Korteweg -- de Vries equation.}
    \label{fig:kdv_time}
\end{figure}

\end{document}